\documentclass[10pt,twocolumn,twoside,final]{IEEEtran}
\pdfoutput=1

\usepackage{array}
\usepackage[dvipdfmx]{graphics}
\usepackage{amsmath,epsfig,graphicx,amsfonts}
\usepackage[hang,raggedright,nooneline]{subfigure}
\usepackage[caption=false,font=footnotesize]{subfig}
\usepackage{subfigmat}
\usepackage{color}
\usepackage{bmpsize}

\begin{document}

\title{A Detail Based Method for Linear Full Reference Image Quality Prediction}

\author{Elio D. Di Claudio\thanks{Copyright \copyright 2010 IEEE. Personal use of this material is permitted. However, permission to use this material for any other purposes must be obtained from the IEEE by sending a request to pubs-permissions@ieee.org.}\thanks{Contact author: Elio D. Di Claudio, Dept. of Information Engineering, Electronics and Telecommunications (DIET), University of Rome ``La Sapienza,'' Via Eudossiana 18, I-00184 Rome, Italy. Ph.: +39-06-44585490, Fax: +39-06-44585632, e-mail elio.diclaudio@diet.uniroma1.it.} and Giovanni Jacovitti\thanks{Giovanni Jacovitti, Dept. of Information Engineering, Electronics and Telecommunications (DIET), University of Rome ``La Sapienza,'' Via Eudossiana 18, I-00184 Rome, Italy. Ph.: +39-06-44585838, Fax: +39-06-44585632, e-mail giovanni.iacovitti@diet.uniroma1.it.}\thanks{EDICS No.: SMR-HPM.}}

\markboth{Submitted to the IEEE Transactions on Image Processing}
{Linear Full Reference Image Quality Prediction}

\maketitle

\begin{abstract}
In this paper, a novel Full Reference method is proposed for image quality assessment, using the combination of two separate metrics to measure the perceptually distinct impact of detail losses and of spurious details. To this purpose, the gradient of the impaired image is locally decomposed as a predicted version of the original gradient, plus a gradient residual. It is assumed that the detail attenuation identifies the detail loss, whereas the gradient residuals describe the spurious details. It turns out that the perceptual impact of detail losses is roughly linear with the loss of the positional Fisher information, while the perceptual impact of the spurious details is roughly proportional to a logarithmic measure of the signal to residual ratio. The affine combination of these two metrics forms a new index strongly correlated with the empirical Differential Mean Opinion Score (DMOS) for a significant class of image impairments, as verified for three independent popular databases. The method allowed alignment and merging of DMOS data coming from these different databases to a common DMOS scale by affine transformations. Unexpectedly, the DMOS scale setting is possible by the analysis of a single image affected by additive noise.
\end{abstract}


\IEEEpeerreviewmaketitle

\section{Introduction}\label{section:Introduction}

\IEEEPARstart{T}{he} interest on the quality assessment of images grows with the volume of visual communications and the advance of imaging technologies, pushing the development of effective techniques for the prediction of the average image quality judged by the human users, hereinafter referred to as the \emph{subjective quality}.

The subjective quality is measured by averaging the scores assigned by a panel of human observers following specific protocols and is usually rated on the so-called Mean Opinion Score (MOS) scale, or on the Differential MOS (DMOS) scale, defined as the MOS difference between the \emph{reference} (original) and the \emph{test} (impaired) images.

However, measurement of subjective quality is unpractical for routine and large scale image quality assessment (IQA). Therefore, the subjective quality is usually predicted by means of \emph{objective} IQA algorithms, founded on abstract models of the human observer.

IQA methods operate in three basic modes. In the Full Reference (FR) mode they quantify the differences between the reference and test images. In the Reduced Reference (RR) mode, the comparison is limited to partial representations of the images. Finally, in the No Reference (NR) mode, selected features extracted from the test image alone are compared to ideal targets \cite{MOORTY11,MITTAL12,SAAD12}.

Surveys of several IQA methods are available in literature (e.g., \cite{CHANDLER13}). Here, to place the presented method in a precise conceptual frame, a concise classification of the IQA methods is provided on the basis of the underlying modeling of the human observer.

\emph{Psycho-physical models} quantify the differences between original and test images by emulating the signal processing of the early stages of the e Human Visual System (HVS) \cite{WINKLER99,CHANDLER07}, supposed typical for all subjects. The generalization capability of these methods is limited by the influence of higher-level mental factors, such as emotion, education, past experience, etc. \cite{CHANDLER06,CAVANAGH11,VU08}.

\emph{Cognitive models} \cite{MARR10} look at the image quality scoring as a \emph{computational process} taking place in the human mind \cite{LARSON10} and assume that the subjective quality degradation is strongly correlated with some measure of \emph{visual information loss}. The Peak Signal to Noise Ratio (PSNR) method is perhaps the simplest cognitive method, which identifies the information loss with a reduction of the signal to noise ratio (SNR), where the difference between the original and the test images is interpreted as noise. 

The Structural Similarity (SSIM) method measures the so-called structural information, defined through the local normalized Mean Square Error (MSE) between the test and the reference images \cite{WANG04}. Variants of the SSIM include Multi-Scale analysis (MS-SSIM) \cite{WANG03} or image gradient analysis, adopted in the Feature Similarity (FSIM) \cite{ZHANG11} and in the Gradient Magnitude Similarity Deviation (GMSD) \cite{XUE14} methods.

The Visual Information (VIF) method measures the loss of Shannon mutual information \cite{SHEIKH06}, while the Virtual Cognitive Method (VICOM) measures the loss of the Fisher information about the localization of patterns \cite{CAPODIFERRO12}. 

As far as cognitive NR methods are concerned, they measure the loss of information as the distance of some statistical features from their nominal values \cite{SIMONCELLI01,ZHAI12}.

The generalization capability of cognitive models depends upon the coverage of adopted information measure. For instance, the PSNR is sensitive to any kind of impairment, whereas the VIF \cite{SHEIKH06} is specialized for impairments modeled by linear distortion plus additive noise.

\emph{Behavioral models} are automatic learning machines trained by samples to give DMOS estimates directly from the images \cite{PINSON04,LI12} or from a set of selected features \cite{LIU17}. If behavioral models are applied for known impairments, they may assume a very simple form \cite{WANG00,SAZZAD08,ZHU12}, but in general they must include an automatic classification of the type of impairment. In these cases, the generalization is a critical issue. Neural nets can tightly fit the DMOS for specific image data sets using many adjustable weights, but the effects of different impairment factors are hard to predict.

\subsection{Shortcomings of existing methods}
\label{sec:ShortcomingsOfExistingMethods}

One typical problem of most FR IQA methods is their unequal sensitivity with respect to the covered impairments \cite{SHEIKH06,CAPODIFERRO12}. To compensate for this drawback, some methods combine values of the same metric calculated at different resolutions \cite{WANG03}, relying on the fact that blur effects vanish when resolution shrinks, whereas noise and artifacts are only attenuated. Another compensation technique exploits the inhomogeneous spatial distribution of errors generated by smoothing and additive effects, measured by the variance of the local MSE \cite{XUE14}.

A more systematic solution to this problem lies in the combination of different metrics specialized for different impairments. This approach was proposed in \cite{MARTENS96}, where two metrics were used to measure the amounts of blur and additive noise. This objective was also pursued in \cite{PINSON04} for video. In \cite{CAPODIFERRO12} a categorical index was paired to the positional Fisher information. Recently, a similar approach was followed in \cite{IEREMEIEV16}.

Another issue overlooked by most existing IQA methods is that the relationship between metrics and the empirical DMOS scale is strongly non-linear \cite{CAPODIFERRO12,GU13}. As a matter of fact, the Spearman Rank Order Correlation Coefficient (SROCC) \cite{HUBER81}, often employed to compare the performance of different metrics, is insensitive to linearity issues. On the other hand, the linearity of the DMOS estimates is essential in applications, since the quality must be quantified at the end on the DMOS scale.

To circumvent this problem, it is customary to \emph{linearize} the metrics versus the empirical DMOS scale using ad hoc non-linear parametric (\emph{logistic}) functions \cite{SHEIKH06B}, as suggested also by the Video Quality Expert Group \cite{VQEG03}. However, unequal sensitivity of metrics to different impairments cannot be compensated by linearization and produces irreducible DMOS fitting errors.

Moreover, the linearization involves the non-linear optimization of several parameters, that critically depend on the empirical data set. The generalization capability across different data sets degrades linearly with the number of adjustable parameters \cite{AKAIKE74,BURNHAM04}, regardless of the training procedure \cite{STONE77}, and may lead to under-fitting or over-fitting effects, strong statistical scattering of sample fitting parameters and local minima issues. 

These are serious obstacles for the effectiveness of IQA metrics. As a matter of fact, default values for linearization parameters are hard to find in the technical literature. In conclusion, the intrinsic linearity of metrics with the DMOS appears as a highly desirable feature for IQA \cite{OKARMA10,SKUROWSKY14}.

\subsection{The proposed approach}
\label{sec:TheProposedApproach}

The objective of the method presented in this paper is to solve the problems of unequal sensitivity to different impairments and of the a posteriori parametric linearization. Generally speaking, the approach followed here consists of the combination of different metrics, tailored to different impairments. Specifically, it stems from the consideration that most existing IQA methods treat image detail loss and spurious details in the same way. In other words, they do not distinguish between impairments caused by the loss of visual structures (\emph{depriving errors}) or by the appearance of artifacts (\emph{meddling errors}). However, this is at odds with common evidence, since detail losses and spurious details have a very different visual appearance.

Therefore, it is reasonable to measure their effects on subjective quality using different metrics. Previous attempts in this direction appeared in \cite{LUBIN97,CAPODIFERRO10B}, where point-wise increments and decrements of the gradient magnitude were discriminated. In \cite{LI11} detail losses and additive impairments were separated using a restored version of the test image as a watershed.

Herein, this discrimination is operated through a Least Squares (LS) decomposition of the gradient field of the test image into two components: a component \emph{linearly predicted} from the gradient field of the original image, and the \emph{residual, unpredictable} gradient, The detail loss is then identified by the attenuation of the predicted gradient into a small observation window. Likewise, the presence of spurious detail is identified by the gradient residual observed into a small window. This modeling is suggested by the orthogonality of the residual gradient with respect to the predicted gradients \cite{HAYKIN96}. 

The second step is the definition of two metrics, representing the \emph{perceptual impact} of detail losses and spurious details respectively, each one as linear as possible with respect to the empirical DMOS scores. To this purpose, the contributions to the overall perceptual impact coming from detail losses are quantified by the loss of the detail positional information \cite{NERI04,CAPODIFERRO12}, proportional to the square root of the gradient energy attenuation into a detail window. The perceptual impact of spurious details is quantified by a logarithmic measure of the ratio between the original gradient energy and the residual gradient energy, pooled over the image.

The metrics are considered as the coordinates of a two-dimensional ($2$-D) space following the VICOM \cite{CAPODIFERRO12} scheme. The VICOM is structured as a set of computational layers, starting from the extraction of gradients from the reference and the test images, up to the computation of multiple features that form the coordinates of the virtual cognitive state space. Each point of this space is then mapped to a corresponding DMOS estimate by a parametric function trained on experimental data \cite{CAPODIFERRO12}. For this reason, the method presented here is referred to as \emph{Detail VICOM} (D-VICOM). As a reference for the reader, a block scheme of the D-VICOM is displayed later in Fig. \ref{fig:DVICOMdfg}. 

At this point, the visual diversity of detail losses and spurious details is again invoked to argue that their individual perceptual impacts will contribute independently to determine the overall subjective quality loss. It follows that the mapping function from the state space to the DMOS estimate boils down to a simple \emph{affine} transformation. 

The effectiveness of this new model was preliminarily tested on the LIVE DBR2 \cite{LIVEDBR2} database, where the method exhibited superior performance with respect to competing methods. Then, it was verified on two other independent databases (TID2008 \cite{PONOMARENKO09} and CSIQ \cite{MADPAGE10}) for the classes of impairment common to the three databases.

The outstanding fitting and cross-fitting of the D-VICOM estimates up to an affine transformation of the DMOS scale reveal that these heterogeneous experimental data are highly consistent under the D-VICOM paradigm. This consistency allowed to align and fuse these data into a unique large database, called the SUPERQUARTET. 

Rather unexpectedly, it is also concluded that it is possible to define a \emph{database independent} D-VICOM (ID-VICOM) quality estimator, specifying only conventional DMOS values of an original image and of an its noisy version.

This paper is organized as follows. In Sect. \ref{section:Detail analysis} the decomposition of the gradient of the test image is described, and the results are illustrated by means of visual examples. In Sect. \ref{section:Linear metrics} the metrics are defined and their affine combination is proposed for DMOS estimation.

In Sect. \ref{section:Experiments and results} the DMOS estimates are statistically tested using the LIVE DBR2 \cite{LIVEDBR2} and then verified with the data contained in the TID2008 \cite{PONOMARENKO09} and the CSIQ \cite{MADPAGE10}. The performance of the linear D-VICOM and of some competing methods after logistic linearization is compared on the SUPERQUARTET. In Sect. \ref{section:DirectDMOSScaleSetting} it is shown that the D-VICOM can be quickly calibrated without training. In Sect. \ref{section:Computational budget} the computational costs are discussed and in Sect. \ref{section:Conclusion} the merits of the D-VICOM are finally underlined.

\section{Detail analysis}\label{section:Detail analysis}

The detail analysis is performed on the luminance components $I\left( {\bf{p}} \right)$ of an impaired test image and $\tilde I\left( {\bf{p}} \right)$ of the reference image, that are gray-scale functions of the generic pixel position ${\bf{p}} = \left( {{x_1},{x_2}} \right)$. 

Image gradients were theoretically supported as a relevant feature for IQA in \cite{CAPODIFERRO12} and adopted by recent methods \cite{XUE14}. For the proposed analysis, it is convenient to consider the \emph{Gaussian smoothed complex gradients} $y\left( {\bf{p}} \right)$ and $\tilde y\left( {\bf{p}} \right)$, extracted by the following operator of scale $s$ and unit energy \cite{CAPODIFERRO12}
\begin{equation}
{h_0}\left( {\bf{p}},s \right) = \frac{1}{{s \sqrt \pi }}{e^{ - \frac{{{x_1}^2 + {x_2}^2}}{{2{s^2}}}}}\sqrt {\frac{{{x_1}^2 + {x_2 }^2}}{{{\sigma ^2}}}} \, {e^{j\arg \left( {{{{x_1 + j x_2}}}} \right)}}
\label{eqn:CGG}
\end{equation}
as
\begin{IEEEeqnarray}{l}
	\tilde y \left( {\bf{p}} \right) = \tilde I\left( {\bf{p}} \right) \star {h_0}\left( {{\bf{p}};s} \right) \nonumber\\
	y\left( {\bf{p}} \right) = I\left( {\bf{p}} \right) \star {h_0}\left( {{\bf{p}};s} \right)
	\label{eqn:imageCGG}
\end{IEEEeqnarray}
where $\star$ denotes $2$-D convolution and $\arg$ is the principal phase angle of the complex argument. For $s = 1$ pixel or slightly more, the frequency response of (\ref{eqn:CGG}) well approximates the Contrast Sensitivity Function (CSF) of the HVS front end \cite{LI11, mannos74}, at nominal viewing distance \cite{CAPODIFERRO12}. The operator \eqref{eqn:CGG} summarizes the horizontal and vertical filters commonly used for gradient approximation \cite{XUE14} and is \emph{steerable}, i.e., has the same frequency response for any pattern orientation \cite{DICLAUDIO10}.

The test image gradient $y\left( {\bf{p}} \right)$ is decomposed for any $\bf p$ as the sum of a \emph{linearly distorted} (smoothed) version $\hat y \left( {\bf{p}} \right)$ of the original gradient $\tilde y \left( {\bf{p}} \right)$ (identified as the \emph{distorted} detail) and of an \emph{unpredictable} gradient error $\nu \left( {\bf{p}} \right)$ (considered as a \emph{spurious} detail): 
\begin{equation}
	y\left( {\bf{p}} \right) = \hat y\left( {\bf{p}} \right) + \nu \left( {\bf{p}} \right) \; .
	\label{eqn:reproduced detail}
\end{equation}

The distorted $\hat y\left( {\bf{p}} \right)$ is further modeled by a linear combination of the original gradient $\tilde y \left( {\bf{p}} \right)$ and of a pair of its directionally filtered versions 
\begin{equation}
	\hat y\left( {\bf{p}} \right) = {b_{\bf{p}}}\left( 0 \right)\tilde y\left( {\bf{p}} \right) + {b_{\bf{p}}}\left( 1 \right){\tilde y_1}\left( {\bf{p}} \right) + {b_{\bf{p}}}\left( 2 \right){\tilde y_2}\left( {\bf{p}} \right)
	\label{eqn:blur operator}
\end{equation}
where ${b_{\bf{p}}}\left( 0 \right)$, ${b_{\bf{p}}}\left( 1 \right)$ and ${b_{\bf{p}}}\left( 2 \right)$ are real valued coefficients depending on the position ${\bf{p}}$, and ${\tilde y_1}\left( {\bf{p}} \right)$ and ${\tilde y_2}\left( {\bf{p}} \right)$ are calculated as
\begin{IEEEeqnarray}{l}
{\tilde y_1}\left( {\bf{p}} \right) = \tilde y\left( {\bf{p}} \right) \star h{}_1\left( {{\bf{p}};s} \right) \nonumber\\
{\tilde y_2}\left( {\bf{p}} \right) = \tilde y\left( {\bf{p}} \right) \star h{}_2\left( {{\bf{p}};s} \right) \;.
	\label{eqn:side_gradients}
\end{IEEEeqnarray}

The impulse responses $h_1\left( {{\bf{p}};s} \right)$ and $h_2\left( {{\bf{p}};s} \right)$ are second order normalized Gaussian derivatives (i.e., Hermite-Gauss) functions of the same scale $s$ as ${h_0}\left( {\bf{p}},s \right)$, defined as
\begin{displaymath}
	{h_1}\left( {{\bf{p}};s} \right) = \frac{{2\frac{{x_1^2}}{{{s^2}}} - 1}}{{s\sqrt {2\pi } }}{e^{ - \frac{{x_1^2}}{{2{s^2}}}}}
	\label{eqn:h_1}
\end{displaymath}
\begin{displaymath}
	{h_2}\left( {{\bf{p}};s} \right) = \frac{{2\frac{{x_2^2}}{{{s^2}}} - 1}}{{s\sqrt {2\pi } }}{e^{ - \frac{{x_2^2}}{{2{s^2}}}}} 
	\label{eqn:h_2}
\end{displaymath} 
and are introduced to adequately model a local astigmatic blur \cite{NERI04,DICLAUDIO10}.

The coefficients ${b_{\bf{p}}}\left( k \right)$ for $k=0,1,2$ are identified by a regularized LS system \cite{golub89}, which minimizes the local error energy
\begin{IEEEeqnarray}{c}
\varepsilon \left( {\bf{p}} \right) = \sum\limits_{\bf{q}} {w{{\left( {\bf{q}} \right)}^2}{{\left| {\nu \left( {{\bf p}+{\bf q}} \right) } \right|}^2}} 
\label{eqn:residual_error}
\end{IEEEeqnarray} 
within a spot centered on the analyzed point $\bf p$. To minimize the interference from adjacent points, the squared errors are weighted by a Gaussian window $w\left( {\bf{q}} \right) \propto {e^{ - \frac{{{{\left| {\bf{q}} \right|}^2}}}{{4 s_w^2}}}}$ of spread $s_w \sqrt{2}$, scaled so that $\sum\limits_{\bf{q}} {w{{\left( {\bf{q}} \right)}^2}} = 1$. In particular, the smallest $s_w = 1$ pixel compatible with the number of estimated parameters within the spot was adopted \footnote{For $s_w = 1$ the effective spot has a diameter larger than four pixels. So more than $18$ effective real valued equations are available for each $\bf p$ to estimate the three real $b_{\bf{p}} \left( k \right)$ plus the local error variance.}. 

The overall cost function is 
\begin{equation}
	J\left( {\bf{p}} \right) = \varepsilon \left( {\bf{p}} \right) + \xi\left[ {\sum\limits_{k = 0}^2 {b_{\bf{p}}^2\left( k \right)} } \right]
	\label{eqn:regularized_LScost}
\end{equation}
where the penalty term $\xi = 1$ was chosen for $256$ gray-level images to regularize \cite{golub89} the sample $b_{\bf{p}} \left( {k} \right)$ if the original gradient magnitude is small, without significantly biasing the LS parameter estimate. The LS solution is computationally efficient and its statistical properties are well known \cite{HAYKIN96,tarpey00} (see also the Appendix).

\begin{figure*}[!tb]
\centering
\includegraphics[width=7in]{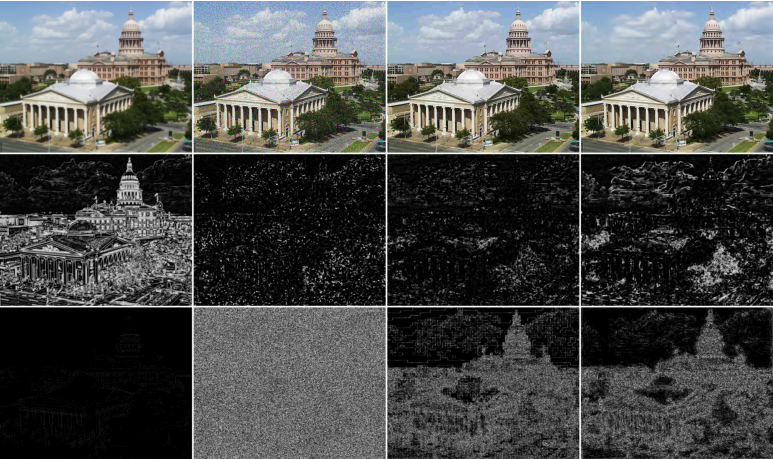}
\caption{First row: blurred (left), noisy, JPEG and JPEG2000 (right) coded versions of the \emph{Church and Capitol} image from the LIVE database \protect\cite{LIVEDBR2}. Second row: gradient attenuation maps of the above images. Third row: gradient residual maps.}
\label{fig1}
\end{figure*}

To put into evidence how the proposed decomposition is actually correlated with the visual findings, in Fig. \ref{fig1} the \emph{gradient attenuation map}
\begin{displaymath}
	l\left( {\bf{p}} \right) = 1 - \frac{{\left| {\hat y\left( {\bf{p}} \right)} \right| + V}}{{\left| {\tilde y\left( {\bf{p}} \right)} \right| + V}}\;,
	\label{eqn:detail_loss}
\end{displaymath}
an indicator of detail loss\footnote{The constant $V = 20$ was inserted for display regularization purposes.}, and the residual gradient magnitude map $\rho\left( {\bf{p}} \right) = \left| {\nu \left( {\bf{p}} \right)} \right|$, an indicator of the spurious detail presence, are displayed for blurred, noisy, JPEG and JPEG2000 compressed images. 

Maps are enhanced to reveal details well below the visibility threshold. The noisy image is characterized by random gradient residuals and, conversely, the blurred image is characterized by diffuse gradient attenuation and negligible residuals. The coded images, that are affected by both lost and spurious details, exhibit mostly complementary patterns of gradient attenuation and gradient residuals.

\section{Metrics}\label{section:Linear metrics}

\subsection{Detail loss metric}\label{section:DetailLossMetric}

The perceptual impact of detail loss was not directly analyzed in the past. The closest problem considered in literature was the FR rating of noiseless images blurred by optical devices, characterized by their Modulation Transfer Function \cite{granger72,carlson80,barten90}.

In this work the impact of the detail loss on the subjective quality rating is analyzed from a different viewpoint. In \cite{CAPODIFERRO12} it was stressed that the subjective degradation of an image should be strictly correlated with the accuracy of pattern \emph{localization}, since it plays an essential role for distance estimation in binocular vision. Since the HVS perception of spatial displacements is basically linear \cite{stevens62}, it is argued that the subjective degradation should be proportional to the increase of the position uncertainty of the observed distorted patterns with respect to the original ones. 

Now, accepting that the performance of vision mechanisms is near optimal, it follows that the visual localization accuracy of a detail is measured by the Fisher information (FI) about its position. In \cite{NERI04,CAPODIFERRO12} the \emph{positional FI} of a portion of image extracted by a window $w{{\left( {\bf{q}} \right)}^2}$ centered on $\bf p$, in the presence of additive Gaussian white noise of variance $\sigma _N^2$, was calculated as
\begin{equation}
	FI({\bf{p}}) = \sigma _N^{ - 2}\left[ {\sum\limits_{\bf{q}} {w{{\left( {\bf{q}} \right)}^2}{{\left| {g\left( {{\bf{p}} + {\bf{q}}} \right)} \right|}^2}} } \right]
	\label{eqn:PFI}
\end{equation}
where $g\left( {\bf{p}} \right)$ is the \emph{true} image gradient. 

It is deduced that the positional accuracies of the original and test images, respectively characterized by the gradients $\tilde g\left( {\bf{p}} \right)$ and $\hat g\left( {\bf{p}} \right)$, are given by
\begin{equation}
\sqrt {\tilde{FI}({\bf{p}})} = \sigma _N^{ - 1}\sqrt {{{\sum\limits_{\bf{q}} {w{{\left( {\bf{q}} \right)}^2}\left| {\tilde g\left( {{\bf{p}} + {\bf{q}}} \right)} \right|} }^2}} 
\label{eqn:PositionalAccuracyOriginal}
\end{equation}
\begin{equation}
\sqrt {\hat{FI}({\bf{p}})} = \sigma _N^{ - 1}\sqrt {{{\sum\limits_{\bf{q}} {w{{\left( {\bf{q}} \right)}^2}\left| {\hat g\left( {{\bf{p}} + {\bf{q}}} \right)} \right|} }^2}} 
\label{eqn:PositionalAccuracyDistorted}
\end{equation}
for the same noise variance $\sigma _N^2$.

Using the \emph{smoothed} gradients, the loss of positional accuracy due to \emph{detail loss only} is estimated as
\begin{equation}
e^{(0)} = \frac{{\sum\limits_{\bf{p}} {\sqrt {\hat \lambda \left( {\bf{p}} \right)} } }}{{\sum\limits_{\bf{p}} {\sqrt {\tilde \lambda \left( {\bf{p}} \right)} } }}
\label{eqn:detaillossquality}
\end{equation}
after posing
\begin{equation}
\tilde \lambda \left( {\bf{p}} \right) = \sum\limits_{\bf{q}} {w{{\left( {\bf{q}} \right)}^2}{{\left| {\tilde y\left( {{\bf{q}} + {\bf{p}}} \right)} \right|}^2}} \label{eqn:tilde_lambda}
\end{equation}
\begin{equation}
\hat \lambda \left( {\bf{p}} \right) = \left[ {\sum\limits_{\bf{q}} {w{{\left( {\bf{q}} \right)}^2}{{\left| {\hat y\left( {{\bf{q}} + {\bf{p}}} \right)} \right|}^2}} } \right] - \alpha \hat \mu \left( {\bf{p}} \right)\label{eqn:hat_lambda}
\end{equation}
where
\begin{equation}
\hat \mu \left( {\bf{p}} \right) = \sum\limits_q {w{{\left( {\bf{q}} \right)}^2}{{\left| {\nu \left( {{\bf{p}} + {\bf{q}}} \right)} \right|}^2}} 
\label{eqn:Mu}
\end{equation}
and the coefficient $\alpha = 0.56$, valid for $s_w = 1$, takes into account the worst case spurious noise prediction in \eqref{eqn:regularized_LScost}, as described in Appendix. The same window $w{{\left( {\bf{q}} \right)}^2}$ is used for \eqref{eqn:regularized_LScost}, \eqref{eqn:PositionalAccuracyOriginal} and \eqref{eqn:PositionalAccuracyDistorted}, so that the effective area interested for detail loss evaluation is about four times the prediction spot, due to the convolutional spread.
 
In \eqref{eqn:detaillossquality} $\hat \lambda \left( {\bf{p}} \right)$ is clipped within the admissible interval $\left[ {0,\tilde \lambda \left( {\bf{p}} \right)} \right]$ to increase its robustness to estimation errors. 

However, expression \eqref{eqn:detaillossquality} has to be refined considering that:
\begin{itemize}
\item the FI \eqref{eqn:PFI} is defined using the \emph{ideal} gradient, which amplifies the spectral components of the image proportionally to the spatial frequency. However, the magnitude response of the smoothed gradient operator \eqref{eqn:CGG} decays at high spatial frequencies. Since the detail loss is mostly characterized by the attenuation of high frequency components, the gradient loss is actually under-estimated by the smoothed gradients \eqref{eqn:tilde_lambda} and \eqref{eqn:hat_lambda}. Therefore, the FI calculated from the smoothed gradient is under-estimated in spots characterized by a rich high frequency content. In this work, the compensation of the FI estimate was obtained in a simple way by raising the magnitude of \eqref{eqn:tilde_lambda} and \eqref{eqn:hat_lambda} to an exponent $\gamma > 1$. The optimal value of $\gamma$ is hard to predict and must be empirically identified. Empirical DMOS values of the three independent databases employed in this work are well explained by setting $\gamma = 1.5$ (see Fig. \ref{fig4});
\item the pooling should be extended only to the set of points ${\bf{p}} \in \left\{ P \right\}$ where the gradient prediction by \eqref{eqn:regularized_LScost} is reliable. In particular, the spots characterized by severe ill-conditioning due to unbalanced energy of the LS equations \cite{HUBER81} are excised from the pooling set. In particular, the spots almost exactly centered on strong edges are discarded, according to the heuristic rule
\begin{equation}
P = \left\{ {{\bf{p}}:\tilde y\left( {\bf{p}} \right) < 0.3\mathop {\max }\limits_{\bf{p}} \left[ {\left| {\tilde y\left( {\bf{p}} \right)} \right|} \right]} \right\} \; .
\label{eqn:poolingset}
\end{equation}
\item To enhance the independence between metrics, full weight should be given only to the points where spurious details are negligible, indicating that the model \eqref{eqn:blur operator} is really accurate, while the weight should be reduced everywhere detail loss overlaps with spurious details.
\end{itemize}

Then, the proposed estimate of the perceptual impact due to details loss only is
\begin{equation}
	e = \frac{{\sum\limits_{{\bf{p}} \in \left\{ P \right\}} {{\rho ^ - }\left( {\bf{p}} \right)\hat \lambda {{\left( {\bf{p}} \right)}^{{\gamma \mathord{\left/
 {\vphantom {\gamma 2}} \right.
 \kern-\nulldelimiterspace} 2}}} + \upsilon } }}{{\sum\limits_{{\bf{p}} \in \left\{ P \right\}} {{\rho ^ - }\left( {\bf{p}} \right)\tilde \lambda {{\left( {\bf{p}} \right)}^{{\gamma \mathord{\left/
 {\vphantom {\gamma 2}} \right.
 \kern-\nulldelimiterspace} 2}}} + \upsilon } }}
	\label{eqn:detaillossfinal}
\end{equation}
where $\upsilon = 0.1$ is a small regularizing constant, $\gamma = 1.5$ and 
\begin{equation}
{\rho ^ - }\left( {\bf{p}} \right) = {\rm{ }}\left\{ {\begin{array}{*{20}{l}}
1&{{\rm{if}}\quad \hat \mu \left( {\bf{p}} \right) < 0.01\tilde \lambda \left( {\bf{p}} \right)\;}\\
{0.25}&{{\rm{elsewhere}}}
\end{array}} \right.
\label{eqn:PoolingWeights}
\end{equation}
is a non critical weighting factor. Finally, the DMOS component attributed to pure detail loss is estimated as
\begin{equation}
{d^ - } = 1 - e \;.
\label{eqn:dminus}
\end{equation}
 
\subsection{Spurious detail metric}\label{section:DetailadditionMetric}

The spurious details are substantially meaningless for the observer, so that their localization accuracy should not be relevant as in the case of detail loss. It is rather plausible that the annoyance produced by spurious details is similar to that caused by random noise. Since for images affected by additive noise the DMOS is quite well estimated by the PSNR logarithmic index, it is assumed that the perceptual impact caused by spurious details is logarithmically related to the ratio between the average energy ${{\tilde \lambda }_{av}}$ of the original details and the average energy ${{\hat \mu }_{av}}$ of the lost details, defined as
\begin{IEEEeqnarray}{rl}
{{\tilde \lambda }_{av}} \,& = \frac{{\sum\limits_{{\bf{p}} \in \left\{ P \right\}} {\tilde \lambda \left( {\bf{p}} \right)} }}{{\sum\limits_{{\bf{p}} \in \left\{ P \right\}} 1 }}\\
{{\hat \mu }_{av}} \,& = \frac{{\sum\limits_{{\bf{p}} \in \left\{ P \right\}} {\hat \mu \left( {\bf{p}} \right)} }}{{\sum\limits_{{\bf{p}} \in \left\{ P \right\}} 1 }} \;.
\label{eqn:average_quantities}
\end{IEEEeqnarray}

Imposing that the subjective quality estimate is zero for diverging noise, let us define the quantity
\begin{equation}
f = \ln \left( {1 + c \frac{{{{\tilde \lambda }_{av}}}}{{{{\hat \mu }_{av} + \sigma _V^2}}}} \right)
\label{eqn:logscore}
\end{equation} 
where the constant $\sigma _V^2 = 20$ accounts for gradient noise variance which starts to impair the quality for white noise corrupted images \cite{SHEIKH06}, and the positive constant $c$ sets the working point of the logarithmic law. It has been found that the empirical DMOS values are well explained by the non-critical value $c = 0.1$ (see Fig. \ref{fig4}).

Finally, the proposed estimate of the perceptual impact attributed to spurious details only is calculated as\footnote{It is worth noting that $\mathop {\lim }\limits_{{{\tilde \lambda }_{av}} \to 0} \left( t \right) = \frac{{\sigma _V^2}}{{{{\hat \mu }_{av} + \sigma _V^2}}}$ is always finite.}
\begin{equation}
	t = \frac{f}{{{f_0}}} = \frac{{\ln \left( {1 + c\frac{{{{\tilde \lambda }_{av}}}}{{{{\hat \mu }_{av} + \sigma _V^2}}}} \right)}}{{\ln \left( {1 + c \frac{{{{\tilde \lambda }_{av}}}}{{\sigma _V^2}}} \right)}} 
	\label{eqn:detailadditionquality}
\end{equation} 
where $t = 1$ is imposed for identical test and reference images. Thus the estimate of the DMOS component attributed to spurious details is
\begin{equation}
	{d^ + } = 1 - t \;.
	\label{eqn:detaillossmetric}
\end{equation}

\subsection{Metric combination}\label{section:MetricCombination}

Following the VICOM scheme \cite{CAPODIFERRO12}, the overall DMOS prediction, referred to as $\mathord{\buildrel{\lower3pt\hbox{$\scriptscriptstyle\smile$}} 
\over D} $, is calculated by combining ${d^ - }$ and ${d^ + }$ through a general \emph{two-dimensional} parametric function 
\begin{equation}
	\mathord{\buildrel{\lower3pt\hbox{$\scriptscriptstyle\smile$}} 
\over D} = f\left( {{d^ - },{d^ + };{\bf{a}}} \right)
\label{eqn:MetricsCombination}
\end{equation}
where $\bf a$ is the fitting coefficient vector. 

The basic assumption of this paper is that detail losses and spurious details are distinctly perceived, as evident by common experience. For a Taylor series expansion of \eqref{eqn:MetricsCombination} around any point $\left( {d_0^ - ,d_0^ + } \right)$, this hypothesis is expressed by
\begin{equation}
\frac{{{\partial ^{m + n}}f\left( {{d^ - },{d^ + };{\bf{a}}} \right)}}{{\partial {{\left( {{d^ - }} \right)}^m}\partial {{\left( {{d^ + }} \right)}^n}}} \approx 0 
\label{eqn:cross_sensitivities}
\end{equation}
for $m,n \ge 1$, leading to the decoupled form
\begin{equation}
	f\left( {{d^ - },{d^ + };{\bf{a}}} \right) = {f^ - }\left( {{d^ - };{{\bf{a}}^ - }} \right) + {f^ + }\left( {{d^ + };{{\bf{a}}^ + }} \right) \;.
	\label{eqn:decoupled_f}
\end{equation}

In addition, if the marginal metrics ${f^ - }\left( {{d^ - };{{\bf{a}}^ - }} \right) $ and ${f^ + }\left( {{d^ + };{{\bf{a}}^ + }} \right) $ are affine functions of ${d^ - }$ and ${d^ + }$, respectively, the approximation \eqref{eqn:MetricsCombination} boils down to
\begin{equation}
	\mathord{\buildrel{\lower3pt\hbox{$\scriptscriptstyle\smile$}} 
\over D} = {a_0} + a_1^ - {d^ - } + a_1^ + {d^ + }
\label{eqn:LinearMetricsCombination}
\end{equation}
where the constant $a_0$ accounts for possible non-zero DMOS scores assigned to the original images during experimental sessions, while $a_1^-$ and $a_1^+$ compensate for the different DMOS sensitivity with respect to ${d^ - }$ and ${d^ + }$. The last point will be discussed in detail in Sect. \ref{subsection:AnInvarianceProperty}.

The form of \eqref{eqn:LinearMetricsCombination} is deemed valid for impairments coming from a mixture of detail loss and spurious detail addition over small spots and distributed all over the test image, as it happens for instance in coding and error correction applications. Other impairments, such as luminance and contrast changes, chromatic aberrations, strong and isolated artifacts and low-frequency, correlated noise, are not directly covered by the present D-VICOM model.

The D-VICOM decomposition scheme is loosely related to the Most Apparent Distortion (MAD) scheme, based on a HVS model \cite{LARSON10}. The basic difference is that the MAD uses two distinct metrics for high and low quality regions (instead of lost and spurious details), according to a visibility map.

\section{Experiments and results}\label{section:Experiments and results}

The linearity and the uniformity of the \emph{scatterplot} of the IQA metric versus the empirical DMOS (see, for instance, Fig. \ref{fig3} in the sequel) of different databases is the ultimate requirement in applications. To achieve this goal, IQA metrics generally need a linearization through a parametric function \cite{SHEIKH06B}. The linearized scatterplot should be tightly concentrated around the diagonal. 

MSE and LCC are good indicators of the scatterplot concentration \cite{SHEIKH06B}, if a wide number of test images, uniformly distributed across the DMOS range, is available, the outliers \cite{HUBER81} of the empirical DMOS have been properly excised and the fitting parameter estimates are not critically influenced by a specific image content \cite{tarpey00}. On the other hand, the proper training of the D-VICOM \eqref{eqn:LinearMetricsCombination} requires the presence in the data set of a class of mixtures covering from essentially spurious detail addition to essentially lost details.

For these reasons, the statistical analysis of the proposed DMOS estimator was performed on large image databases, containing test images affected by different impairments. Specifically, the LIVE DBR2 \cite{LIVEDBR2}, the TID2008 \cite{PONOMARENKO09} and the CSIQ \cite{LARSON10} databases were chosen. These independent databases contain annotated DMOS values obtained through different experimental settings and share four common impairment classes: additive white noise, Gaussian blur, JPEG and JPEG2000 coding. 

The statistical performance of D-VICOM and other competing estimators is illustrated in the sequel, listing the number of the adjustable parameters of the DMOS fitting function. Except for the classical VICOM \cite{CAPODIFERRO12}, linearization was performed by a five parameter monotonic logistic function \cite{SHEIKH06B,VQEG03}, trained by the BRLS modified Newton algorithm \cite{parisi96} over several runs, to minimize the risk of trapping into local minima.

Ordinary LS fitting was used to directly optimize the RMSE. Since we are mainly interested in the linearization parameters, it is worth noting that ordinary LS fitting coefficients are unbiased under the reasonable assumptions of zero mean empirical DMOS distribution, even in the case of different variance across test images \cite{HAYKIN96}.

All metrics were computed by on-line available software \cite{LIVEDBR2,fsimpage13,gmsdpage14,MADPAGE10,msssimpage11}. A preliminary statistical analysis was performed on the full LIVE DBR2 \cite{LIVEDBR2}, which includes impairment classes \cite{SHEIKH06} all covered by the D-VICOM model. Then, the statistical analysis was focused on the classes of impairment common to all three databases. 

\subsection{LIVE DBR2 analysis}\label{subsection:LIVEDBR2Analysis}

LS fitting of \eqref{eqn:LinearMetricsCombination} onto the empirical LIVE DBR2 \cite{LIVEDBR2} DMOS scores yields the following estimator:
\begin{equation}
	\mathord{\buildrel{\lower3pt\hbox{$\scriptscriptstyle\smile$}} 
\over D} = 7.8 + 71. 2\, {d^ - } + 47.0 \,{d^ + } \;.
\label{eqn:LIVEDVICOM}
\end{equation}

\begin{figure}[!tb]
\centering
\includegraphics[width=3.2in]{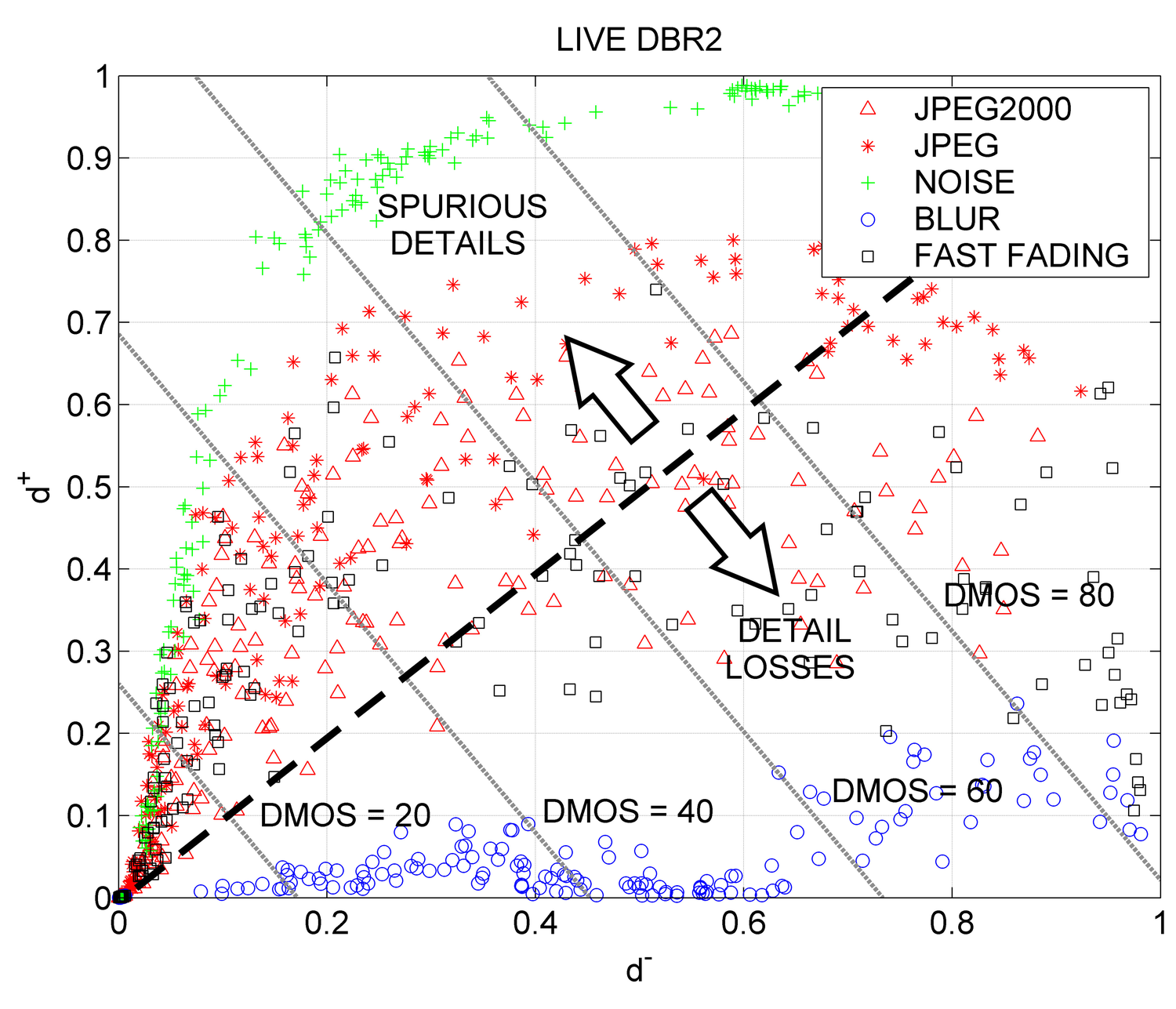}
\caption{The cognitive chart for the LIVE DBR2 \cite{LIVEDBR2}. The iso-metric lines for the D-VICOM regression \eqref{eqn:LIVEDVICOM} are superimposed.}
\label{fig2}
\end{figure}

Following the VICOM scheme \cite{CAPODIFERRO12}, the \emph{virtual cognitive state} of each distorted image is defined by the pair $\left( {d^ -, d^ + }\right)$. The two-dimensional mapping of the cognitive states constitutes the \emph{cognitive chart}, depicted in Fig. \ref{fig2} for the LIVE DBR2 \cite{LIVEDBR2}. The relationship between the cognitive state associated to each test image and the DMOS estimated by \eqref{eqn:LIVEDVICOM} can be read on the superimposed iso-metric DMOS lines, which are rectilinear. They provide an immediate visualization of the quality level of the whole data set. 

As expected, blurred images, mostly characterized by lost details, and noisy images, mostly characterized by spurious details, cluster in the proximity of the coordinate axes. This empirical verification supports the fundamental assumption made in this paper about the distinct perception of detail loss and spurious details. 

Nevertheless, at very high noise levels, gray-scale saturation induces detail loss, and, for very strong blur, halos generate spurious details. A nice surprise is that the metrics $d ^ -$ and $d ^ +$ remain admirably complementary, even in these extreme regions.

JPEG and JPEG2000 images cluster into distinct chart regions, pointing out the different management of detail losses and artifacts operated by these compression techniques \cite{CAPODIFERRO12}. Fast fading states reflect a prevalence of spurious details for light impairments and of detail loss for strong impairments. 

In Table \ref{table:VICOM statistics} the RMSE, the SROCC and the LCC of the D-VICOM are compared to the ones of popular methods, after logistic linearization \cite{SHEIKH06B}. 
\begin{table}[htb]
\caption{SROCC, RMSE and LCC of existing quality estimators and D-VICOM for the LIVE DBR2.}
\centering
\renewcommand{\arraystretch}{1.}
\setlength\tabcolsep{6pt}
\begin{tabular}{@{}|c|c|c|c|c|}
\hline\noalign{\smallskip}
Model & Parameter no. & RMSE & SROCC & LCC\\
\noalign{\smallskip}
\hline
\noalign{\smallskip}
PSNR & $5$ & $13.40$ & $0.876$ & $0.872$ \\
MS-SSIM & $5$ & $8.638$ & $0.951$ & $0.949$ \\
VIF & $5$ & $7.853$ & $0.964$ & $0.958$ \\
VICOM & $5$ & $6.686$ & $0.970$ & $0.969$ \\
FSIM & $5$ & $7.693$ & $0.963$ & $0.960$ \\
MAD & $5$ & $6.802$ & $0.968$ & $0.969$ \\
GMSD & $5$ & $7.625$ & $0.963$ & $0.960$ \\
\noalign{\smallskip}
\hline
\noalign{\smallskip}
D-VICOM & $\bf 3$ & $\bf{6.371}$ & $\bf{0.975}$ & $\bf{0.972}$ \\
\noalign{\smallskip}
\hline
\noalign{\smallskip}
\end{tabular}
\label{table:VICOM statistics}
\end{table}

\begin{figure}[!tb]
\centering
\includegraphics[width=3.2in]{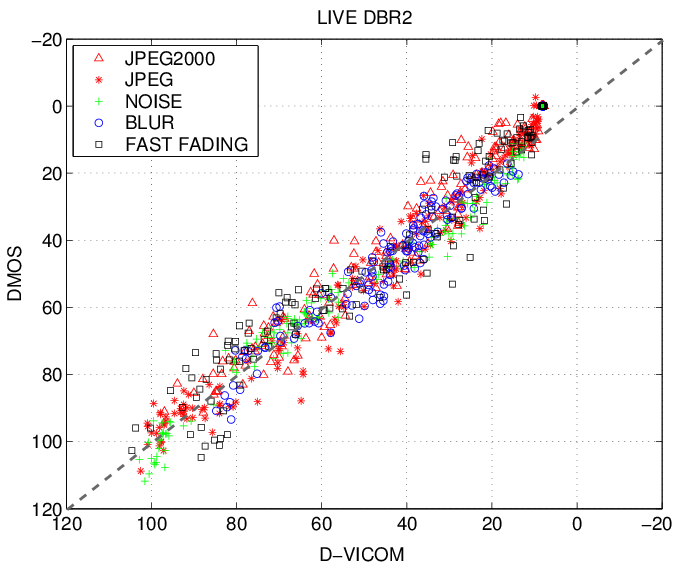}
\includegraphics[width=3.2in]{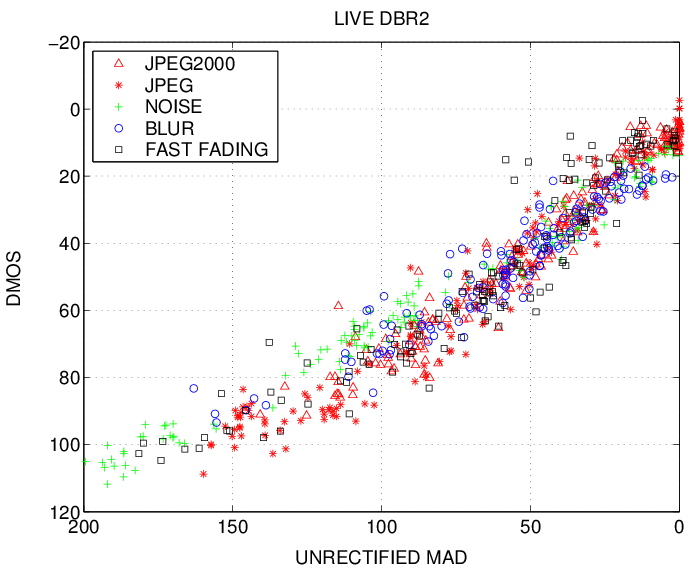}
\includegraphics[width=3.2in]{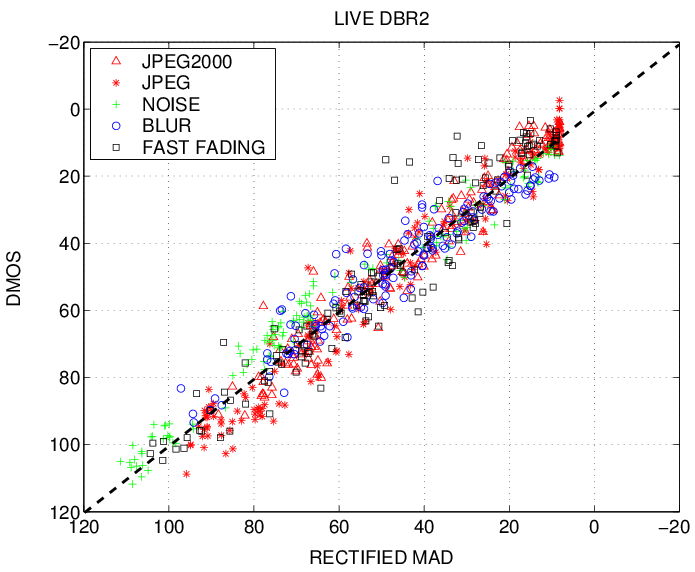}
\caption{Top row: Scatterplot of the D-VICOM estimate versus DMOS for the LIVE DBR2. $RMSE = 6.371$, $SROCC = 0.975$, $LCC = 0.972$. Middle row: Scatterplot of the MAD \protect\cite{LARSON10} scores versus DMOS for the LIVE DBR2 before logistic linearization. Bottom row: Scatterplot of the MAD estimate versus DMOS after logistic linearization. $RMSE = 6.802$, $SROCC = 0.968$, $LCC = 0.969$.}
\label{fig3}
\end{figure}

In Fig. \ref{fig3} the scatterplot of the D-VICOM is compared to the scatterplot of the seemingly closest uni-dimensional competitor, i.e., the MAD estimator \cite{LARSON10}, before and after logistic linearization. The linearity of the D-VICOM is comparable to the linearity of the MAD after logistic transformation. The DMOS scale goes beyond the conventional upper limit of $100$, because the LIVE DBR2 scores were \emph{realigned} among different impairments \cite{SHEIKH06B,LIVEDBR2}.

\begin{figure}[!tb]
\centering
\includegraphics[width=3.2in]{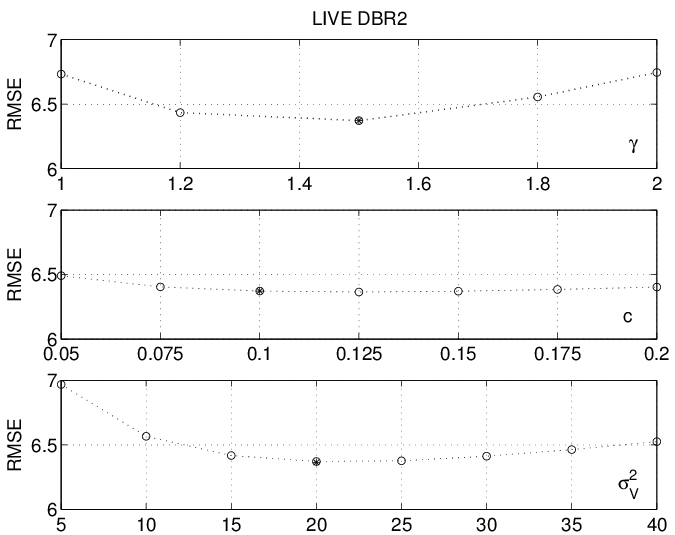}
\caption{Plots showing the negligible marginal sensitivity of the RMSE of the LIVE DBR2 DMOS estimates with respect to the values of (from top to bottom) $\gamma$, $c$ and $\sigma_V^2$, posed as constants in the D-VICOM model in Sect. \ref{section:Linear metrics}. In each plot, the other two quantities are kept fixed at the nominal values ($\gamma = 1.5$, $c = 0.1$ and $\sigma_V^2 = 20$). }
\label{fig4}
\end{figure}

Fig. \ref{fig4} puts into evidence the negligible marginal sensitivity of the D-VICOM RMSE with respect to $\gamma$, $c$ and $\sigma_V^2$ around the values ($\gamma = 1.5$, $c = 0.1$ and $\sigma_V^2 = 20$) adopted as constants in our model.

\subsection{Performance on equivalent subsets of TID2008 and CSIQ databases}
\label{subsection:Performance on equivalent subsets of TID2008 and CSIQ databases}

To validate previous results, the same analysis was extended to the similar subsets of the TID2008 and CSIQ databases. In particular, the five image subsets of TID2008 (white Gaussian noise, Gaussian blur, JPEG and JPEG2000 coding and JPEG2000 transmission errors) already used in \cite{CAPODIFERRO12} and four subsets (white Gaussian noise, Gaussian blur, JPEG and JPEG2000) of the CSIQ were chosen. Fitting results are summarized in Tables \ref{table:TID2008fivesetstatistics} and \ref{table:CSIQfoursetstatistics}. TID2008 MOS scores were translated into DMOS scores for uniformity.

\begin{table}[!hbt]
\caption{SROCC, RMSE, LCC and LCC obtained from the LIVE DBR2 regression coefficients of existing IQA metrics and D-VICOM for the Gaussian noise, Gaussian blur, JPEG and JPEG2000 coding and JPEG2000 transmission errors image subsets of the TID2008 database.}
\centering
\renewcommand{\arraystretch}{1.}
\setlength\tabcolsep{6pt}
\begin{tabular}{@{}|c|c|c|c|c|c|}
\hline\noalign{\smallskip}
Model & Parameter & RMSE & SROCC & LCC & LCC \\
& no. & & & & (LIVE DBR2)\\
\noalign{\smallskip}
\hline
\noalign{\smallskip}
PSNR & $5$ & $9.774$ & $0.840$ & $0.801$ & $0.809$\\
MS-SSIM & $5$ & $7.895$ & $0.893$ & $0.876$ & $0.858$\\
VIF & $5$ & $6.800$ & $0.908$ & $0.910$ & $0.917$\\
VICOM & $5$ & $6.436$ & $0.929$ & $0.919$ & $0.876$\\
FSIM & $5$ & $6.673$ & $0.938$ & $0.913$ & $0.905$\\
MAD & $5$ & $6.005$ & $0.921$ & $0.930$ & $0.914$\\
GMSD & $5$ & $7.200$ & $0.932$ & $ 0.898$ & $0.894$\\
\noalign{\smallskip}
\hline
\noalign{\smallskip}
D-VICOM & $\bf 3$ & $\bf 5.279$ & $\bf 0.956$ & $\bf 0.946$ & $\bf 0.942$\\
\noalign{\smallskip}
\hline
\noalign{\smallskip}
\end{tabular}
\label{table:TID2008fivesetstatistics}
\vspace{12 pt}
\caption{SROCC, RMSE, LCC and LCC obtained from the LIVE DBR2 regression coefficients of existing quality estimators and D-VICOM for the Gaussian noise, Gaussian blur, JPEG and JPEG2000 coding image subsets of the CSIQ database.}
\centering
\renewcommand{\arraystretch}{1.}
\setlength\tabcolsep{6pt}
\begin{tabular}{@{}|c|c|c|c|c|c|}
\hline\noalign{\smallskip}
Model & Parameter & RMSE & SROCC & LCC & LCC\\
& no. & & & & (LIVE DBR2)\\
\noalign{\smallskip}
\hline
\noalign{\smallskip}
PSNR & $5$ & $0.1466$ & $0.922$ & $0.855$ & $0.891$\\
MS-SSIM & $5$ & $0.0915$ & $0.953$ & $0.946$ & $0.952$\\
VIF & $5$ & $0.0722$ & $0.919$ & $0.924$ & $0.960$\\
FSIM & $5$ & $0.1017$ & $0.962$ & $0.934$ & $0.963$\\
MAD & $5$ & $0.0667$ & $0.967$ & $ 0.972$ & $\bf 0.973$\\
GMSD & $5$ & $0.0701$ & $0.969$ & $0.969$ & $0.969$\\
\noalign{\smallskip}
\hline
\noalign{\smallskip}
D-VICOM & $\bf 3$ & $\bf 0.0646$ & $\bf 0.972$ & $\bf 0.974$ & $ \bf 0.973$\\
\noalign{\smallskip}
\hline
\noalign{\smallskip}
\end{tabular}
\label{table:CSIQfoursetstatistics}
\end{table}

\begin{figure*}[!tb]
	\centering
	\includegraphics[width=6.4in]{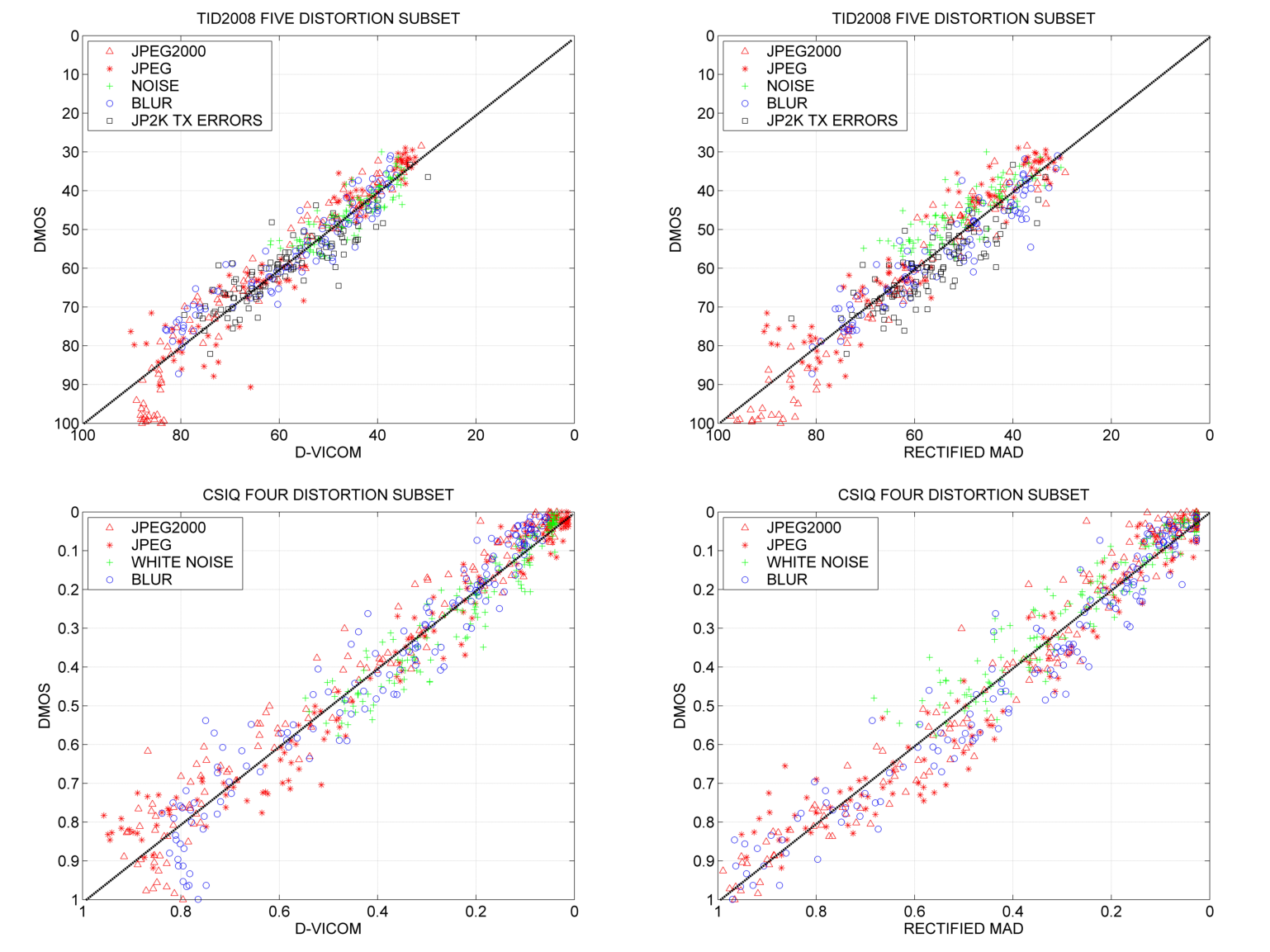}	
	\caption{First row: (left) Scatterplot of the D-VICOM versus DMOS for the five distortion TID2008 subset, $RMSE = 5.279$, $SROCC = 0.956$, $LCC = 0.946$; (right) Scatterplot of the linearized MAD versus DMOS for the five distortion TID2008 subset, $RMSE = 6.005$, $SROCC = 0.921$, $LCC = 0.930$. Second row: (left) Scatterplot of the D-VICOM versus DMOS for the four distortion CSIQ subset, $RMSE = 0.0646$, $SROCC = 0.972$, $LCC = 0.974$; (right) Scatterplot of the linearized MAD versus DMOS for the CSIQ four distortion subset, $RMSE = 0.0667$, $SROCC = 0.967$, $LCC = 0.972$. }
	\label{fig8}
\end{figure*}

For these subsets, the accuracy of the D-VICOM is still the best one. This is also visible from the comparison between the scatterplots of the D-VICOM and of the linearized MAD (the seemingly most performing competitor) reported in Fig. \ref{fig8}.

\subsection{Statistical analysis}
\label{subsection:Statistical analysis}

Returning back to the Tables \ref{table:TID2008fivesetstatistics} and \ref{table:CSIQfoursetstatistics}, the rightmost column contains the LCC obtained by applying the regression coefficients computed for the the largest and most complete data set in our analysis (the LIVE DBR2) on the target subset to assess the \emph{generalization} capabilities of IQA metrics. Considering that the LCC is insensitive to affine metric and/or DMOS transformations\footnote{The SROCC magnitude of uni-dimensional metrics is invariant for monotonic transformations. The SROCC is a \emph{robust} version of the LCC \cite{HUBER81} and therefore it is equally insensitive to true outliers, non-linear fitting and not well explained scores.}, results support the generalization power of good IQA metrics and, at the same time, show that D-VICOM is uniformly the most performing subjective quality predictor in this \emph{cross-validation} exercise, achieving very close results between the two fitting scenarios. 

Moderate non Gaussianity and local additive effects of DMOS errors are expected features \cite{SHEIKH06B}, due to the low number of human observers involved in subjective tests and the generally non-robust \cite{HUBER81} protocols for score averaging and outlier rejection. Therefore it is useful to check the leave-one-out cross validation (LOOCV) RMSE, i.e., the square root of the \emph{Predicted Sum of Squares} (PRESS) \cite{tarpey00}, scaled by the subset size, since it measures the \emph{expected RMSE} for a test image added to the data set. The LOOCV RMSE is \emph{analytically} computable for D-VICOM, depending on the projection (\emph{hat}) matrix \cite{HUBER81,tarpey00} of the LS system matrix generated by \eqref{eqn:LinearMetricsCombination}, and yields $6.395$ for the LIVE DRB2, $5.316$ for the TID2008 subset and $0.0650$ for the CSIQ subset. All these values are \emph{extremely} close to the corresponding fitting RMSE, confirming the completeness and the DMOS stability of tried subsets, as well as the good conditioning of the LS fittings.

For the other metrics, the empirical LOOCV RMSE computation is unpractical, but, assuming Gaussianity and homoscedasticity of the DMOS errors within each subset, it can be estimated for large samples by the Akaike Information Criterion (AIC) \cite{AKAIKE74,STONE77}, which can be simplified for the generic $k$th IQA metric as
\begin{equation}
	AIC_k = 2N \ln \left({RMSE_k}\right) + 2 \left({P_k+1}\right)
	\label{eqn:aic}
\end{equation}
for a subset size $N$ and $P_k$ adjustable parameters\footnote{The RMSE is a further nuisance parameter for LS regression.}. AIC application does not require nested models \cite{STONE77}, is summable across databases for overall evaluation and admits a significance test through the analysis of the (raw) AIC weights \cite{BURNHAM04} ${\Delta _k} = {e^{ - 0.5\left[ {AI{C_k} - \mathop {\min }\limits_l \left( {AI{C_l}} \right)} \right]}} $. In particular, $\Delta_k < 0.1$ (i.e., AIC scores different by more than $4$ nats) indicate high confidence ($\approx 95\%$) in the model with the smallest AIC \cite{BURNHAM04}.

For the three analyzed subsets, the AIC and ${\Delta_k}$ scores are listed in Table \ref{table:DBR2TID5CSIQ4statistics2}. The AIC significance test is passed by the D-VICOM in all cases. In the same table, the residual kurtosis and the $95$th percentile ($95p$) of residual magnitudes provide additional insight about the fitting error properties. 

However, the kurtosis and $95p$ values do not indicate clear differences or worrisome issues across metrics and subsets, with a moderate non Gaussianity of the residuals and an acceptable $p95 \approx 2{\mkern 1mu} \cdot RMSE$. 

On the other hand, the small typical LCC ($<0.5$) between the residual vectors of the best metrics here (the D-VICOM and the MAD) suggests that there is a substantial excess variance not explained by the models. So a comparison of the LS error variances by F-test for \emph{independent} Gaussian variables \cite{snedecor89} is useful. 

In this case, the $95\%$ one-sided confidence level about the significance of the D-VICOM RMSE improvement is attained whenever the RMSE of a competing IQA metric exceeds $6.767$ for the LIVE DBR2 database, $5.684$ for the TID2008 subset and $0.0691$ for the CSIQ subset. This confidence level is always attained by the D-VICOM, except against the classical VICOM for the LIVE DBR2 and the MAD for the CSIQ subset.

\begin{table*}[!hbt]
\caption{Residual kurtosis, $95p$, AIC and raw AIC weights $\Delta_k$ of existing IQA metrics and D-VICOM for the LIVE DBR2 database, the five distortion TID2008 subset (TID5) and the four distortion CSIQ (CSIQ4) subset.}
\centering
\renewcommand{\arraystretch}{1.}
\setlength\tabcolsep{6pt}
\begin{tabular}{@{}|c|c|c|c|c|c|c|c|c|c|c|c|c|}
\hline\noalign{\smallskip}
& \multicolumn{3}{c|}{Kurtosis} & \multicolumn{3}{c|}{$95p$} & \multicolumn{3}{c|}{AIC} & \multicolumn{3}{c|}{$\Delta_k$} \\
\noalign{\smallskip}
\hline\noalign{\smallskip}
Model & LIVE & TID5 & CSIQ4 & LIVE & TID5 & CSIQ4 & LIVE & TID5 & CSIQ4 & LIVE & TID5 & CSIQ4 \\
\noalign{\smallskip}
\hline\noalign{\smallskip}
PSNR & $3.06$ & $4.37$ & $3.48$ & $27.12$ & $21.48$ & $0.28$ & $4051.7$ & $2291.7$ & $-2292.1$ & $0.00$ & $0.00$ & $0.00$ \\ 
MS-SSIM & $3.74$ & $2.95$ & $4.32$ & $17.33$ & $15.24$ & $0.18$ & $3371.3$ & $2078.2$ & $-2857.7$ & $0.00$ & $0.00$ & $0.00$ \\ 
VIF & $2.83$ & $3.47$ & $5.84$ & $15.17$ & $13.74$ & $0.15$ & $3222.9$ & $1928.9$ & $-3142.0$ & $0.00$ & $0.00$ & $0.00$ \\ 
FSIM & $4.47$ & $3.99$ & $3.58$ & $15.25$ & $13.56$ & $0.19$ & $3190.8$ & $1910.1$ & $-2730.9$ & $0.00$ & $0.00$ & $0.00$ \\ 
MAD & $4.29$ & $3.16$ & $3.67$ & $13.94$ & $12.36$ & $0.13$ & $2999.0$ & $1804.6$ & $-3237.1$ & $0.00$ & $0.00$ & $0.00$ \\ 
GMSD & $5.35$ & $4.38$ & $4.78$ & $15.35$ & $15.96$ & $0.15$ & $3177.0$ & $1986.1$ & $-3177.4$ & $0.00$ & $0.00$ & $0.00$ \\ 
\noalign{\smallskip}
\hline
\noalign{\smallskip}
D-VICOM & $3.49$ & $4.22$ & $3.90$ & $12.84$ & $10.90$ & $0.13$ & $2893.0$ & $1671.7$ & $-3279.4$ & $1.00$ & $1.00$ & $1.00$ \\ 
\noalign{\smallskip}
\hline
\noalign{\smallskip}
\end{tabular}
\label{table:DBR2TID5CSIQ4statistics2}
\end{table*}

\subsection{Un-covered impairments}
\label{subsection:Un-covered impairments}

To meet the possible, legitimate curiosity of the reader, in Tables \ref{table:TID2008statistics} and \ref{table:CSIQstatistics} the statistical performance indexes are reported for the whole TID2008 and CSIQ archives, that contain impairments not explicitly modeled by the D-VICOM scheme. The average accuracy of D-VICOM is still acceptable, even if theoretically unsupported.

\begin{table}[!hbt]
\caption{SROCC, RMSE and LCC of existing quality estimators and D-VICOM for the entire TID2008 database.}
\centering
\renewcommand{\arraystretch}{1.}
\setlength\tabcolsep{6pt}
\begin{tabular}{@{}|c|c|c|c|c|}
\hline\noalign{\smallskip}
Model & Parameter no. & RMSE & SROCC & LCC\\
\noalign{\smallskip}
\hline
\noalign{\smallskip}
PSNR & $5$ & $12.74$ & $0.553$ & $0.519$ \\
MS-SSIM & $5$ & $7.968$ & $0.854$ & $0.845$ \\
VIF & $5$ & $9.169$ & $0.750$ & $0.789$ \\
FSIM & $5$ & $7.256$ & $0.881$ & $0.874$ \\
MAD & $5$ & $8.278$ & $0.835$ & $0.832$ \\
GMSD & $5$ & $\bf 7.156$ & $\bf 0.891$ & $\bf 0.877$ \\
\noalign{\smallskip}
\hline
\noalign{\smallskip}
D-VICOM & $\bf 3$ & $7.923$ & $0.834$ & $0.837$ \\
\noalign{\smallskip}
\hline
\noalign{\smallskip}
\end{tabular}
\label{table:TID2008statistics}
\vspace{12 pt}
\caption{SROCC, RMSE, LCC of existing quality estimators and D-VICOM for the entire CSIQ database.}
\centering
\renewcommand{\arraystretch}{1.}
\setlength\tabcolsep{6pt}
\begin{tabular}{@{}|c|c|c|c|c|}
\hline\noalign{\smallskip}
Model & Parameter no. & RMSE & SROCC & LCC\\
\noalign{\smallskip}
\hline
\noalign{\smallskip}
PSNR & $5$ & $0.1719$ & $0.806$ & $0.756$ \\
MS-SSIM & $5$ & $0.1275$ & $0.910$ & $0.874$ \\
VIF & $5$ & $0.1004$ & $0.919$ & $0.924$ \\
FSIM & $5$ & $0.1197$ & $0.922$ & $0.890$ \\
MAD & $5$ & $\bf 0.0804$ & $0.949$ & $\bf 0.952$ \\
GMSD & $5$ & $0.0824$ & $\bf 0.957$ & $0.949$ \\
\noalign{\smallskip}
\hline
\noalign{\smallskip}
D-VICOM & $\bf 3$ & $0.0961$ & $0.937$ & $0.931$ \\
\noalign{\smallskip}
\hline
\noalign{\smallskip}
\end{tabular}
\label{table:CSIQstatistics}
\end{table}

\subsection{An invariance property}
\label{subsection:AnInvarianceProperty}

Consistency aspects of human perception of quality among different people are relevant for IQA, as discussed in \cite{wajid14}. The analysis performed so far neglected the relative perceptual sensitivity of lost and spurious details on the DMOS. In the sequel, an invariance property is deduced from the following \emph{equivalence argument}: a subject has to indicate the blur strength which yields the same subjective quality loss of a fixed amount of noise added to the same image. There is not any apparent reason why the response should change among different people with normal vision capability under the same viewing conditions, at least in an average sense.

In a formal setting, it is assumed that two generic databases $A$ and $B$, characterized by the D-VICOM coefficient sets \eqref{eqn:LinearMetricsCombination} $\left\{ {{a_{0A}},a_{1A}^ - ,a_{1A}^ + } \right\}$ and $\left\{ {{a_{0B}},a_{1B}^ - ,a_{1B}^ + } \right\}$, share the same original images and that a generic blurred image contained in $A$ and $B$ has DMOS $\mathord{\buildrel{\lower3pt\hbox{$\scriptscriptstyle\smile$}} \over D}_A $ and $\mathord{\buildrel{\lower3pt\hbox{$\scriptscriptstyle\smile$}} \over D}_B $ respectively. Since blur essentially does not add spurious details, $d ^ + \approx 0$. 

Moreover, let us imagine that in both databases there exists an \emph{equivalent} image contaminated by additive noise and characterized by the same DMOS values $\mathord{\buildrel{\lower3pt\hbox{$\scriptscriptstyle\smile$}} \over D}_A $ and $\mathord{\buildrel{\lower3pt\hbox{$\scriptscriptstyle\smile$}} \over D}_B $. Since white noise essentially does not introduce detail loss, $d ^ - \approx 0$.

Inserting these values into \eqref{eqn:LinearMetricsCombination} yields
\begin{IEEEeqnarray}{rl}
\mathord{\buildrel{\lower3pt\hbox{$\scriptscriptstyle\smile$}} \over D}_A = {a_{0A}} + a_{1A} ^ - d ^ - & = {a_{0A}} + a_{1A}^+ d ^ + \\
\mathord{\buildrel{\lower3pt\hbox{$\scriptscriptstyle\smile$}} \over D}_B = {a_{0B}} + a_{1B} ^ - d ^ - & = {a_{0B}} + a_{1B}^+ d ^ + \;.
\label{eqn:blurnoise}
\end {IEEEeqnarray}

Simplifying, it turns out that the ratio
\begin{equation}
	r = \frac{{{d^ + }}}{{{d^ - }}} = \frac{{a_{1A}^ - }}{{a_{1A}^ + }} = \frac{{a_{1B}^ - }}{{a_{1B}^ + }}
	\label{eqn:UniversalRatio}
\end{equation}
should not change across databases. On the basis of this equivalence argument, the D-VICOM is rewritten as
\begin{equation}
	\mathord{\buildrel{\lower3pt\hbox{$\scriptscriptstyle\smile$}} 
\over D} = {a_0} + a_1^ + \left( {{d^ + } + r{d^ - }} \right)
\label{eqn:D-VICOMtwinpar}
\end{equation}
where $r$ is kept as a \emph{constant}. This argument can be extended to the actual cases where images are different among databases, stating that the D-VICOM estimator \eqref{eqn:D-VICOMtwinpar} can be tuned to any empirical database by adjusting only the pair $\left( {{a_0},a_1^ + } \right)$, i.e., the estimator \emph{offset} and \emph{slope}. This result confirms at a glance the excellent D-VICOM LCC cross-validation results of Sect. \ref{subsection:Performance on equivalent subsets of TID2008 and CSIQ databases}. 

Statistical identification of $r$ started from a coarse guess obtained by the regression coefficients \eqref{eqn:LinearMetricsCombination} of each \emph{quartet} of distortions common to LIVE DBR2, TID2008 and CSIQ databases. Then an iterative LS optimization of the pairs $\left( {{a_0},a_1^ + } \right)$ of each quartet in \eqref{eqn:D-VICOMtwinpar} and of a unique $r$ on the \emph{joint} database (seven parameters in total) was performed by minimizing the sum of the squared DMOS residuals of the three subsets\footnote{CSIQ DMOS values were up-scaled by $100$ to balance residual energies.}. The cost function achieved a broad minimum at $r=1.64$, so that the D-VICOM estimator \eqref{eqn:D-VICOMtwinpar} assumes the \emph{operative} form 
\begin{equation}
	\mathord{\buildrel{\lower3pt\hbox{$\scriptscriptstyle\smile$}} 
\over D} = {a_0} + a_1^ + \left( {{d^ + } + 1.64{d^ - }} \right)
\label{eqn:ID-VICOMtwinpar}
\end{equation} 
characterized by only \emph{two} parameters, offset ($a_0$) and slope ($a ^ + $), to be tuned to the desired DMOS scale. Table \ref{table:TwoparDVICOMstatistics} shows that the statistical performance of the two-parameter D-VICOM remains substantially unchanged, as expected. 

\begin{figure}
	\centering
		\includegraphics[width=2.8in]{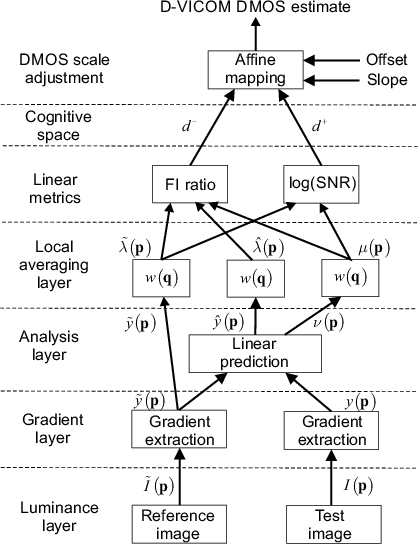}
	\caption{The flow chart of the D-VICOM which shows the computational layers.}
	\label{fig:DVICOMdfg}
\end{figure}

The overall flow chart of the D-VICOM \eqref{eqn:ID-VICOMtwinpar} is shown in Fig. \ref{fig:DVICOMdfg}. 

To explore the impairment coverage of \eqref{eqn:ID-VICOMtwinpar}, Table \ref{table:TwoparDVICOMstatisticsTIDdetail} lists the statistical indexes of the D-VICOM, the MAD, and the GMSD (i.e., the most robust metric from Table \ref{table:CSIQstatistics}) on the $17$ image subsets of the TID2008 \cite{PONOMARENKO09}, trained on the five impairment subset previously used, deemed \emph{safe} for ordinary LS fitting. 

The RMSE strongly varies among subsets and metrics, complicating the suitability check of IQA metrics on specific impairment sets. However, AIC minimization, under the proper statistical assumptions, remains useful for a quick significance test \cite{BURNHAM04}. Assuming in this scenario zero mean Gaussian fitting errors with a different $RMSE_{mk}$ for the $k$th metric and the $m$th subset, containing $N_m$ images, and $P_k$ fitting parameters, an upper bound to the $AIC_k$ for the wider set $M = \left\{ {{m_1},{m_2}, \ldots } \right\}$ is:
\begin {displaymath}
AI{C_k} \le \sum\limits_{m \in M} 2{\left[ {{N_m}\ln \left( {RMS{E_{mk}}} \right) + 1} \right]} + 2{P_k} \;.
\label{eqn:compoundAIC}
\end{displaymath}

The AIC verification effort is still combinatorial, but from Table \ref{table:TwoparDVICOMstatisticsTIDdetail} the D-VICOM \eqref{eqn:ID-VICOMtwinpar} emerges as the best overall choice as regards the fitting RMSE (and therefore the AIC), except for low frequency impairments (quantization noise, block distortions, contrast and intensity changes), barely observable by the gradient operator \eqref{eqn:CGG}, but exhibits a surprising robustness to impulse distortion. 

\begin{table}[htb]
\caption{SROCC, RMSE and LCC of the two parameter D-VICOM \eqref{eqn:ID-VICOMtwinpar} on various image subsets.}
\centering
\renewcommand{\arraystretch}{1.}
\setlength\tabcolsep{6pt}
\begin{tabular}{@{}|c|c|c|c|c|}
\hline\noalign{\smallskip}
Subset & RMSE & SROCC & LCC & AIC \\
\noalign{\smallskip}
\hline
\noalign{\smallskip}
LIVE DBR2 (full) & $6.415$ & $0.975$ & $0.972$ & $2901.8$ \\
TID2008 (5 dist. subset) & $5.344$ & $0.955$ & $0.945$ & $1682.0$\\
TID2008 (full) & $7.946$ & $0.830$ & $0.846$ & $7053.1$ \\
CSIQ (4 dist. subset) & $0.0648$ & $0.973$ & $0.973$ & $-3277.7$ \\
CSIQ (full) & $0.1065$ & $0.918$ & $0.914$ & $-3783.0$ \\
\hline
\noalign{\smallskip}
\end{tabular}
\label{table:TwoparDVICOMstatistics}
\end{table}

\begin{table*}[htb]
\centering
\renewcommand{\arraystretch}{1.}
\setlength\tabcolsep{6pt}
\caption{SROCC, RMSE and LCC of D-VICOM \eqref{eqn:ID-VICOMtwinpar}, MAD, and GMSD on TID2008 image subsets after LS fitting on the five distortion set.}
\label{table:TwoparDVICOMstatisticsTIDdetail}
\begin{tabular}{|c|c|c|c|c|c|c|c|c|c|c|}
\hline\noalign{\smallskip}
 \multicolumn{2}{|c|}{Model} & \multicolumn{3}{c|}{D-VICOM \eqref{eqn:ID-VICOMtwinpar}} & \multicolumn{3}{c|}{MAD} & \multicolumn{3}{c|}{GMSD} \\ 
\noalign{\smallskip}
\hline\noalign{\smallskip}
No. & Subset name \cite{PONOMARENKO09} & RMSE & SROCC & LCC & RMSE & SROCC & LCC & RMSE & SROCC & LCC \\
\noalign{\smallskip}
\hline\noalign{\smallskip}
1 & Additive Gaussian noise & $\bf 3.971$ & $0.877$ & $0.856$ & $6.380$ & $0.838$ & $0.816$ & $4.053$ & $\bf 0.918$ & $\bf 0.897$ \\ 
2 & Additive noise in color... & $3.394$ & $0.884$ & $0.884$ & $3.881$ & $0.827$ & $0.823$ & $\bf 2.882$ & $\bf 0.898$ & $\bf 0.896$ \\ 
3 & Spatially correlated noise & $\bf 5.069$ & $0.887$ & $0.877$ & $5.659$ & $0.868$ & $0.860$ & $6.960$ & $\bf 0.913$ & $\bf 0.916$ \\ 
4 & Masked noise & $\bf 4.925$ & $\bf 0.851$ & $\bf 0.867$ & $5.094$ & $0.734$ & $0.760$ & $6.072$ & $0.709$ & $0.569$ \\ 
5 & High frequency noise & $\bf 4.266$ & $0.907$ & $\bf 0.936$ & $8.530$ & $0.887$ & $0.894$ & $4.448$ & $\bf 0.919$ & $0.928$ \\ 
6 & Impulse noise & $\bf 7.562$ & $\bf 0.823$ & $\bf 0.802$ & $14.28$ & $0.065$ & $0.042$ & $7.602$ & $0.661$ & $0.627$ \\ 
7 & Quantization noise & $10.627$ & $0.838$ & $0.786$ & $13.29$ & $0.819$ & $0.800$ & $\bf 4.536$ & $\bf 0.890$ & $\bf 0.882$ \\ 
8 & Gaussian blur & $\bf 4.710$ & $\bf 0.954$ & $\bf 0.944$ & $5.116$ & $0.926$ & $0.934$ & $6.161$ & $0.897$ & $0.894$ \\ 
9 & Image denoising & $6.243$ & $0.948$ & $0.955$ & $\bf 5.687$ & $0.943$ & $0.961$ & $6.332$ & $\bf 0.975$ & $\bf 0.978$ \\ 
10 & JPEG compression & $5.651$ & $0.935$ & $0.956$ & $6.444$ & $0.927$ & $0.949$ & $\bf 4.629$ & $\bf 0.952$ & $\bf 0.985$ \\ 
11 & JPEG2000 compression & $6.734$ & $0.969$ & $0.964$ & $\bf 5.540$ & $0.971$ & $0.974$ & $10.30$ & $\bf 0.980$ & $\bf 0.978$ \\ 
12 & JPEG transmission errors & $9.911$ & $\bf 0.867$ & $\bf 0.860$ & $\bf 7.766$ & $0.863$ & $0.854$ & $9.402$ & $0.862$ & $0.852$ \\ 
13 & JPEG2000 transmission errors & $\bf 5.252$ & $0.874$ & $\bf 0.875$ & $6.417$ & $0.839$ & $0.830$ & $8.793$ & $\bf 0.883$ & $0.868$ \\ 
14 & Non eccentricity pattern noise & $\bf 8.859$ & $0.746$ & $0.742$ & $13.71$ & $\bf 0.829$ & $\bf 0.825$ & $12.65$ & $0.760$ & $0.756$ \\ 
15 & Local block-wise distortions... & $17.61$ & $0.640$ & $0.650$ & $11.07$ & $0.798$ & $0.801$ & $\bf 3.880$ & $\bf 0.897$ & $\bf 0.900$ \\ 
16 & Mean shift (intensity shift) & $8.642$ & $0.475$ & $0.529$ & $8.789$ & $0.521$ & $0.576$ & $\bf 5.160$ & $\bf 0.650$ & $\bf 0.677$ \\ 
17 & Contrast change & $13.30$ & $\bf 0.676$ & $\bf 0.727$ & $19.75$ & $0.239$ & $0.215$ & $\bf 11.87$ & $0.465$ & $0.542$ \\ 
\noalign{\smallskip}
\hline\noalign{\smallskip}
\multicolumn{2}{|c|}{Training set (subsets $\left\{ {1,8,10,11,13} \right\}$)} & $\bf 5.344$ & $\bf 0.955$ & $\bf 0.945$ & $6.005$ & $0.921$ & $0.930$ &$7.200$ &$0.932$ & $0.898$ \\
\hline\noalign{\smallskip}
\multicolumn{2}{|c|}{Overall} & $8.296$ & $0.830$ & $0.846$ & $9.631$ & $0.835$ & $0.830$ & $\bf 7.366$ & $\bf 0.891$ & $\bf 0.877$ \\
\noalign{\smallskip}
\hline
\end{tabular}
\end{table*}
	
\subsection{The SUPERQUARTET}
\label{subsection:TheSUPERQUARTET}

The empirical DMOS values of the TID2008 and of the CISQ databases were mapped onto the LIVE DBR2 DMOS scale using the slope/offset pairs estimated in the previous step, forming a unique large database called the \emph{SUPERQUARTET}\footnote{A similar realignment of different databases was already performed in \cite{wolf02} for video sequences, on a space of seven features. The entire SUPERQUARTET realigned DMOS database is freely available upon request to the contact author. It contains $634$ impaired images from the LIVE DBR2, $400$ from the TID2008 and $600$ from the CSIQ, for a total of $1634$ images.}, viewed as the LIVE DBR2 populated by compatible samples migrated from the TID2008 and the CSIQ databases, after proper realignment of the DMOS values to cope with protocol differences. 

\begin{figure}[!tb]
\centering
\includegraphics[width=3.2in]{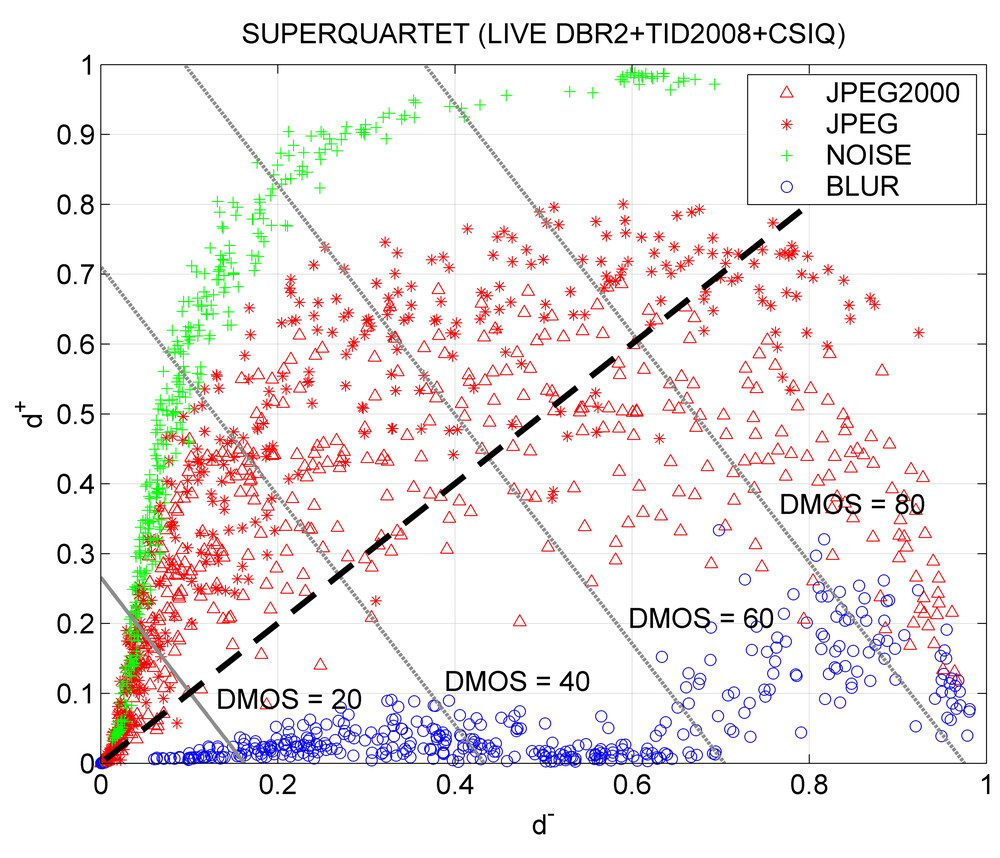}
\caption{The cognitive chart for the SUPERQUARTET. The iso-metric lines for the D-VICOM regression \eqref{eqn:SuperquartetDVICOM} are superimposed.}
\label{fig5}
\end{figure}

\begin{figure}[!tb]
\centering
\includegraphics[width=3.2in]{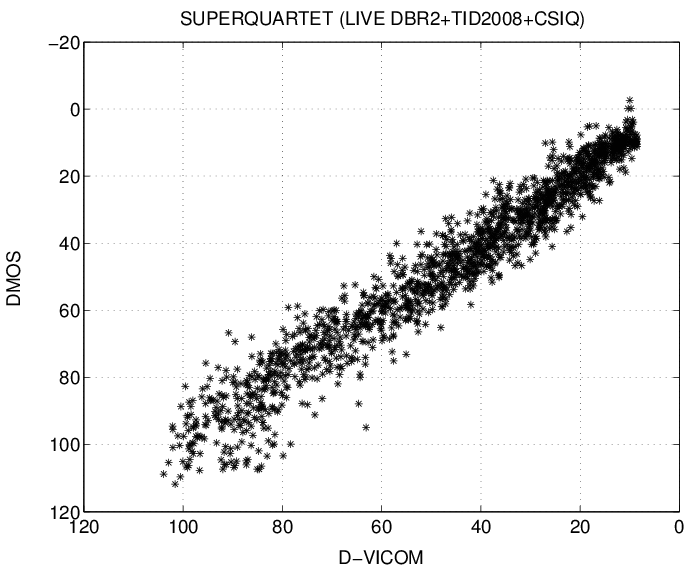}
\centering
\includegraphics[width=3.2in]{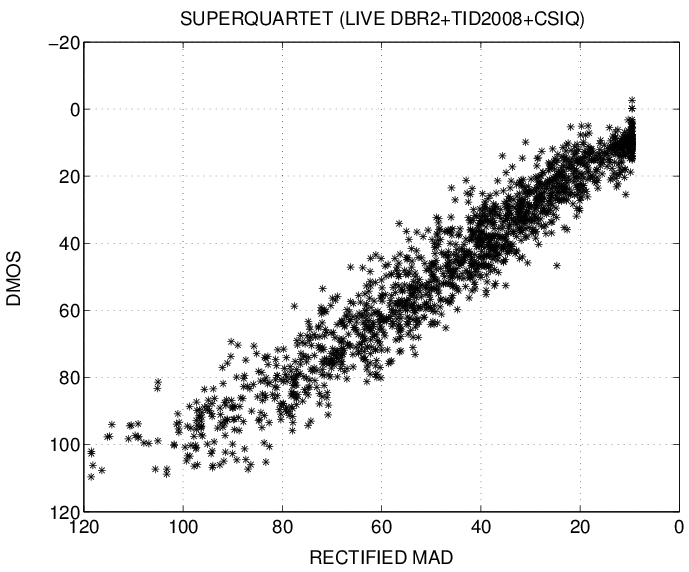}
\centering
\caption{Top: The two parameter D-VICOM scatterplot of the SUPERQUARTET, realigned on the LIVE DBR2 scale. $RMSE = 6.299$, $SROCC = 0.975$ and $LCC = 0.971$; Bottom: The MAD scatterplot over the SUPERQUARTET after the linearization by the five parameter logistic. $RMSE = 7.305$, $SROCC = 0.966$ and $LCC = 0.962$. }
\label{fig6}
\end{figure}

The D-VICOM and a set of popular FR quality estimators, linearized by a logistic function \cite{SHEIKH06B}, were applied to the SUPERQUARTET. Since the DMOS scale is unique (i.e., the LIVE DBR2 one), all estimators were tuned using a single set of parameters (five for the linearization of uni-dimensional IQA estimators and one offset and slope pair for the D-VICOM). Table \ref{table:SUPERQUARTETstatistics} lists the RMSE, the SROCC, the LCC and theAIC scores. In Fig. \ref{fig5} the cognitive chart of the SUPERQUARTET is displayed. The state clusters of all databases remarkably \ overlap. In Fig. \ref{fig6} the scatterplots of the SUPERQUARTET DMOS values versus the optimal estimates of D-VICOM and MAD \cite{LARSON10}, the seemingly best competitor from Table \ref{table:SUPERQUARTETstatistics}, are shown. 

\begin{table}[htb]
\caption{The SROCC, the RMSE, the LCC and the AIC of existing quality estimators and D-VICOM for the SUPERQUARTET.}
\centering
\renewcommand{\arraystretch}{1.}
\setlength\tabcolsep{6pt}
\begin{tabular}{@{}|c|c|c|c|c|c|}
\hline\noalign{\smallskip}
Model & Parameter no. & RMSE & SROCC & LCC & AIC\\
\noalign{\smallskip}
\hline
\noalign{\smallskip}
PSNR & $5$ & $14.34$ & $0.892$ & $0.840$ & $8714.9$\\
MS-SSIM & $5$ & $8.984$ & $0.949$ & $0.941$ & $7186.7$\\
VIF & $5$ & $8.052$ & $0.960$ & $0.953$ & $6828.8$\\
FSIM & $5$ & $7.555$ & $0.966$ & $0.958$ & $6620.6$\\
MAD & $5$ & $7.305$ & $0.966$ & $0.962$ & $6510.6$\\
GMSD & $5$ & $7.627$ & $0.968$ & $0.958$ & $6651.6$\\
\noalign{\smallskip}
\hline
\noalign{\smallskip}
D-VICOM & $2$ & $\bf{6.299}$ & $\bf{0.975}$ & $\bf{0.971}$ & $\bf 6020.4$ \\
\noalign{\smallskip}
\hline
\noalign{\smallskip}
\end{tabular}
\label{table:SUPERQUARTETstatistics}
\end{table}

In turn, the excellent cross-fitting of the LIVE DBR2, the TID2008 and the CSIQ data, up to a simple affine transformation, demonstrates the remarkable statistical homogeneity of these independent experiments under the D-VICOM paradigm. 

In conclusion, the following quality estimator
\begin{equation}
{\mathord{\buildrel{\lower3pt\hbox{$\scriptscriptstyle\smile$}} 
\over D} } = 8.0 + 45.0 \left( {{d^ + } + 1.64{d^ - }} \right)
\label{eqn:SuperquartetDVICOM}
\end{equation}
is proposed as a conventional \emph{calibration free} metric, adjusted to the LIVE DBR2 DMOS scale for the covered impairments. This estimator, hereinafter called \emph{ID-VICOM}, when applied to the \emph{entire} LIVE DBR2, yielded $RMSE = 6.431$, $SROCC = 0.975$ and LCC = $0.972$, nearly identical to the top performance of the three-parameter D-VICOM in Table \ref{table:VICOM statistics}, as well as $RMSE = 6.050$, $SROCC = 0.977$ and $LCC = 0.974$ on the joint TID+CSIQ realigned quartets.

\section{Direct DMOS scale setting}
\label{section:DirectDMOSScaleSetting}

However, the LIVE DBR2 DMOS scale adopted for ID-VICOM might not be adequate in some applications. It is possible to set any desired scale by conveniently specifying another \emph{offset} and \emph{slope} pair, using a \emph{single} test image. To this purpose, the offset of the DMOS scale is fixed to the desired value for perfect images\footnote{The sample DMOS is usually affected by equivocation between original and slightly distorted images.}, say ${a_0} = {a_{0U}}$, so that 
\begin{equation}
	{\mathord{\buildrel{\lower3pt\hbox{$\scriptscriptstyle\smile$}} 
\over D}} \approx {a_{0U}} + a_{1U}^ + \left( {{d^ + } + 1.64{d^ - }} \right) \;.
\label{eqn:conventionalIDVicom}
\end{equation}

\begin{figure}[!tb]
	\centering
 \includegraphics[width=3.2in]{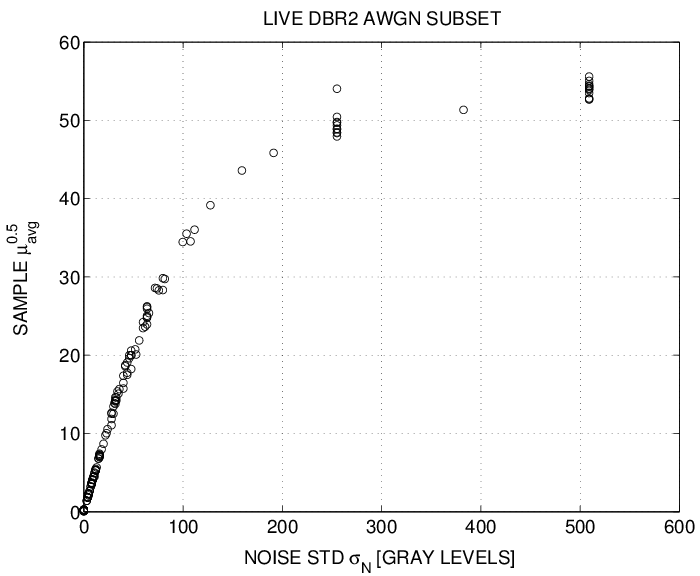}
	\caption{Plot of the white noise standard deviation $\sigma_N$ versus $\sqrt{\hat \mu _{av}}$ for the LIVE DBR2 white noise image subset.}
	\label{fig9}
\end{figure}
 
The slope $a_{1U}^ + $ can be determined from a single image $I_N$ affected by additive white noise of known variance $\sigma_N^2$. In fact, from the diagram of Fig. \ref{fig9}, where the values of $\sqrt{\hat \mu _{av}}$ are plotted versus $\sigma_N$ for the LIVE DBR2 \cite{LIVEDBR2} noisy images, we see that the value of ${\hat \mu _{av}}$ is reliably linked to the noise variance. Moreover ${\tilde \lambda _{av}}$ is calculated from the noise free image. Therefore, $d ^ +$ is obtained from \eqref{eqn:detailadditionquality} and \eqref{eqn:detaillossmetric}. Assigning now a specific DMOS value to $I_N$, the value of $d ^ -$ is read on the cognitive chart of Fig. \ref{fig5} and $a_{1U}^ + $ is finally determined from \eqref{eqn:conventionalIDVicom}.
 
Such a huge generalization power of ID-VICOM \eqref{eqn:conventionalIDVicom} starting from the quality of a single image is surprising at glance, but it is a necessary consequence of the following concurrent facts:
\begin{itemize}
	\item the pronounced linearity of the metrics with the target impairments;
	\item the statistical stability of the additive white noise variance estimate through ${\hat \mu _{av}}$;
	\item the precision of the locus traced by images affected by white noise in the cognitive chart.
\end{itemize}

\section{Computational budget}\label{section:Computational budget}

The major computational burden of D-VICOM lies in the gradient extraction and in the solution of \eqref{eqn:regularized_LScost}, which amounts to $15$ real-valued $2$-D fast correlations/convolution across the image, plus the solution of a linear system of normal equations \cite{golub89} of size three for each image point. With a negligible performance cost, it is also possible to solve \eqref{eqn:regularized_LScost} on a point grid decimated by two on each axis and to linearly interpolate the computed coefficient sets $\left\{ {{b_{\bf{p}}}\left( k \right)} \right\}$ in the remaining points, thus reducing the prediction cost by about $75 \%$.

The average computational time of the D-VICOM MATLAB code with $s = s_w = 1$, running on PC equipped by an Intel Core i7-3370K $3.5$ GHz CPU, was $12.5$ s ($4.7$ s for the fast version\footnote{The basic and fast versions of the D-VICOM are freely available upon request from the contact author.}) on a $1024 \times 768$ pixel image. For the sake of comparison, the MAD ran at the same $12.5$ s, the VIF at $5.1$ s, the FSIM at $1.34$ s, the MS-SSIM at $0.96$ s and the GMSD at only $60$ ms.

\section{Conclusion}\label{section:Conclusion}

The intricate problem of forcing a unique metric to linearly fit subjective scores caused by heterogeneous impairments was circumvented by decomposing it into two easier problems of determining partial metrics linearly related to two well discernible error causes: detail loss and spurious detail addition.

On the basis of the approximate linearity of metrics and of the assumed perceptual properties, it is shown that it is possible to fit these metrics to subjective DMOS ratings by adjusting just the two parameters of an affine transformation. This makes possible to simply merge empirical results coming from different databases for the class of impairments covered by the D-VICOM model.

Last but not least, it is possible to determine a conventional ID-VICOM index, after fixing the DMOS value for one original image and a noisy version of the same image. 

In conclusion, the main merits of the D-VICOM estimator are:
\begin{itemize}
	\item \emph{a priori} connotation of the impairments covered by the method;
	\item outstanding accuracy over a wide quality range;
	\item simple and accurate merging of empirical DMOS from different databases;
	\item analytic diagnostic capabilities through the maps of the gradient attenuation, of the gradient residuals and of the cognitive states;
	\item choice of fast calibration based on a single sample image.
\end{itemize}

Let us finally remark that the D-VICOM quality predictor is not prone to the natural image hypothesis, which supports many existing methods. Moreover, the D-VICOM could be extended to other classes of impairments by adding appropriate metrics. In particular, the D-VICOM is suitable for detail displacement measurements, required in video and $3$-D quality assessment.

\appendices
\section*{Appendix: Noise effects compensation in \eqref{eqn:regularized_LScost}}\label{appendix:NoiseEffectsCompensation}

The LS system \eqref{eqn:regularized_LScost} which computes the estimate ${{\bf{\hat b}}_{\bf{p}}}$ of ${{\bf{ b}}_{\bf{p}}}$ at each point $\bf p$ has the general form:

\begin{equation}
\left[ {\begin{array}{*{20}{c}}
{{\bf{W}}{{{\bf{\tilde Y}}}_r}}\\
{{\bf{W}}{{{\bf{\tilde Y}}}_i}}\\
{\sqrt C {{\bf{I}}_3}}
\end{array}} \right]{{\bf{\hat b}}_{\bf{p}}} \cong \left[ {\begin{array}{*{20}{c}}
{{\bf{W}}{{\bf{y}}_r}}\\
{{\bf{W}}{{\bf{y}}_i}}\\
{\bf{0}}
\end{array}} \right] \approx \left[ {\begin{array}{*{20}{c}}
{{\bf{W}}\left( {{{{\bf{\tilde Y}}}_r}{{\bf{b}}_{\bf{p}}} + {{\bf{n}}_r}} \right)}\\
{{\bf{W}}\left( {{{{\bf{\tilde Y}}}_i}{{\bf{b}}_{\bf{p}}} + {{\bf{n}}_i}} \right)}\\
{\bf{0}}
\end{array}} \right]
\label{eqn:LSSystem}
\end{equation}
where $ {\bf W} $ is a diagonal matrix holding the analysis window weights $w \left({\bf q}\right) $, ${\bf{\tilde Y}}_r$ and ${\bf{\tilde Y}}_i$ are matrices built from the real and the imaginary parts of (filtered) original gradients, ${\bf{\tilde y}}_r$ and ${\bf{\tilde y}}_i$ are vectors built from the real and the imaginary parts of the test image gradients, herein assumed corrupted by the additive noise vectors ${\bf{n}}_r$ and ${\bf{n}}_i$ having zero mean, i.i.d. entries of variance ${{\sigma _N^2} \mathord{\left/ {\vphantom {{\sigma _N^2} 2}} \right. \kern-\nulldelimiterspace} 2}$.

Under above hypotheses and neglecting the small influence of the regularizing constant $\xi$, the predicted target ${\bf{\hat y}}$ can be written as \cite{HAYKIN96} 
\begin{equation}
	{\bf{\hat y}} \approx {\bf{A}}{{\bf{A}}^T}\left[ {\begin{array}{*{20}{c}}
{{\bf{W}}\left( {{{\bf{Y}}_r}{{\bf{b}}_{\bf{P}}} + {{\bf{n}}_r}} \right)}\\
{{\bf{W}}\left( {{{\bf{Y}}_i}{{\bf{b}}_{\bf{P}}} + {{\bf{n}}_i}} \right)}\\
{\bf{0}}
\end{array}} \right]
\label{eqn:predictedimage}
\end{equation}
where $\bf A$ is a three-column orthogonal matrix with the same span as the system matrix. Therefore some noise is added by ${{\bf{\hat b}}_{\bf{p}}}$ to the true predicted gradient with variance 
\begin{equation}
{\hat \sigma ^2} = \frac{{\sigma _N^2}}{2}{\mathop{\rm trace}\nolimits} \left( {{{\bf{A}}^T}\left[ {\begin{array}{*{20}{c}}
{{{\bf{W}}^2}}&{\bf{0}}&{\bf{0}}\\
{\bf{0}}&{{{\bf{W}}^2}}&{\bf{0}}\\
{\bf{0}}&{\bf{0}}&{\bf{0}}
\end{array}} \right]{\bf{A}}} \right)
\label{eqn:PredictedNoise}
\end{equation}
which unduly increases the estimated $\hat \lambda \left( {\bf{p}} \right)$ in \eqref{eqn:hat_lambda}. 

Under the same assumptions, the expected residual energy $E\left[ {J} \right]$ of \eqref{eqn:LSSystem} is
\begin{equation}
	E\left[ {J} \right] = \sigma _N^2 \rm{trace}\left( {{{\bf{W}}^2}} \right) - {\hat \sigma ^2} \; .
	\label{eqn:ResidualVariance}
\end{equation}

The exact calculus of ${\hat \sigma ^2}$ at each $\bf p$ would require a costly full QR solution of \eqref{eqn:regularized_LScost} \cite{golub89}. Following a worst case approach and setting $\rm{trace}\left( {{{\bf{W}}^2}} \right) = 1$, as done throughout the paper, the bound ${\hat \sigma ^2} \le \sigma _N^2{\rm{ }}\left( {w_1^2 + {{w_2^2} \mathord{\left/ {\vphantom {{w_2^2} 2}} \right. \kern-\nulldelimiterspace} 2}} \right)$ holds by the Courant Fischer min-max theorem \cite{golub89}, where $w_1$ and $w_2$ respectively are the largest and the second largest samples of $w \left({\bf q}\right) $. Combining this bound with \eqref{eqn:ResidualVariance}, we get
\begin{equation}
{\hat \sigma ^2} \le \frac{{w_1^2 + {{w_2^2} \mathord{\left/
 {\vphantom {{w_2^2} 2}} \right.
 \kern-\nulldelimiterspace} 2}}}{{1 - \left( {w_1^2 + {{w_2^2} \mathord{\left/
 {\vphantom {{w_2^2} 2}} \right.
 \kern-\nulldelimiterspace} 2}} \right)}}E\left[ J \right]{\rm{ }}\mathop {\rm{ = }}\limits^{def} \alpha E\left[ J \right] 
\label{eqn:AlphaCalculus}
\end{equation}
whose estimate $\alpha \mu \left({\bf p}\right)$ is subtracted from the energy of the predicted gradient in \eqref{eqn:hat_lambda}.
 
This worst case assumption is justified by the fact that most Gaussian-windowed gradient energy is concentrated in few equations of \eqref{eqn:LSSystem}. The efficacy of the proposed correction is demonstrated by the statistical stability of $d ^ -$ for the white noise corrupted images in Figs. \ref{fig2} and \ref{fig5}, at the cost of a negligible positive bias.

\bibliographystyle{IEEEtran}
\bibliography{IEEEabrv,imageprocessing,arrayprocessing}

\end{document}